%% file: preprint.tex
\PassOptionsToPackage{table}{xcolor}
\RequirePackage{fix-cm}
\documentclass{article} 
\usepackage{iclr2027_conference,times}

\input{math_commands.tex}

\usepackage{hyperref}
\usepackage{url}
\usepackage[utf8]{inputenc}
\usepackage[T1]{fontenc}
\usepackage{graphicx}
\usepackage{amsmath}
\usepackage{amssymb}
\usepackage{booktabs}
\usepackage{multirow}
\usepackage{makecell}
\usepackage{array}
\usepackage{tabularx}
\usepackage{longtable}
\usepackage{xcolor}
\usepackage{ifthen}
\usepackage{wrapfig}
\usepackage{capt-of}
\usepackage{placeins}

\providecommand{\paperversion}{arxiv}

\newlength{\halffloatwidth}
\makeatletter
\let\WALA@makecaption\@makecaption
\long\def\@makecaption#1#2{%
  \begingroup
  \def\WALA@tabletype{table}%
  \ifx\@captype\WALA@tabletype
    \setlength{\abovecaptionskip}{0pt}%
    \setlength{\belowcaptionskip}{6pt}%
  \else
    \setlength{\abovecaptionskip}{6pt}%
    \setlength{\belowcaptionskip}{0pt}%
  \fi
  \WALA@makecaption{#1}{#2}%
  \endgroup
}
\makeatother

\definecolor{WALABlue}{RGB}{232,241,252}
\newcommand{\best}[1]{\textbf{#1}}
\newcommand{\second}[1]{\underline{#1}}
\newcommand{\cmark}{\checkmark}
\newcommand{\xmark}{--}

\newcommand{\walaRobotwinClean}{90.62}
\newcommand{\walaRobotwinRandom}{92.80}
\newcommand{\walaRobotwinAvg}{91.71}
\newcommand{\walaWanRobotwinClean}{92.36}
\newcommand{\walaWanRobotwinRandom}{92.30}
\newcommand{\walaWanRobotwinAvg}{92.33}

\title{WALA: Learning Executable Latent Actions from Action-Labeled
Demonstrations and Action-Free Videos}

\newcommand{\WALAauthor}[2]{\mbox{#1\textsuperscript{\fontsize{8}{8}\selectfont#2}}}
\author{%
\begin{tabular}{@{}c@{}}
\WALAauthor{Jiahao Liu}{1,2,3}\hspace{1em}%
\WALAauthor{Yupeng Zheng}{1,2}\hspace{1em}%
\WALAauthor{Zhongpu Xia}{2,\textdagger}\hspace{1em}%
\WALAauthor{Shuai Tian}{1}\\[2pt]
\WALAauthor{Huangrui Li}{1}\hspace{1em}%
\WALAauthor{Yuhang Zheng}{4}\hspace{1em}%
\WALAauthor{Ning Ma}{5}\hspace{1em}%
\WALAauthor{ShangQing Zhou}{1}\\[2pt]
\WALAauthor{Xiaotian Liu}{2}\hspace{1em}%
\WALAauthor{Jing Li}{2}\hspace{1em}%
\WALAauthor{Yixian Li}{2}\hspace{1em}%
\WALAauthor{Haoran Li}{1,3,*}\hspace{1em}%
\WALAauthor{Dongbin Zhao}{1,3}\\[8pt]
{\normalfont\fontsize{10}{12}\selectfont
\textsuperscript{1}SKL-MAIS, Institute of Automation, Chinese Academy of Sciences}\\[2pt]
{\normalfont\fontsize{10}{12}\selectfont
\textsuperscript{2}Anyverse Dynamics}\\[2pt]
{\normalfont\fontsize{10}{12}\selectfont
\textsuperscript{3}School of Advanced Interdisciplinary Sciences, University of Chinese Academy of Sciences}\\[2pt]
{\normalfont\fontsize{10}{12}\selectfont
\textsuperscript{4}National University of Singapore\hspace{1.5em}%
\textsuperscript{5}Xi'an Jiaotong-Liverpool University}\\[5pt]
{\normalfont\fontsize{9}{11}\selectfont
\textsuperscript{*}Corresponding author\hspace{1.5em}\textsuperscript{\textdagger}Project leader}
\end{tabular}%
}

\makeatletter
\newcommand{\WALApreprinttitle}{%
  \begingroup
  \centering
  {\LARGE\scshape
   WALA: Learning Executable Latent Actions\\
   from Action-Labeled Demonstrations\\
   and Action-Free Videos\par}
  \vspace{12pt}
  {\normalfont\fontsize{11}{14}\selectfont\@author\par}
  \endgroup
  \vspace{10pt}
}
\makeatother

\ifthenelse{\equal{\paperversion}{anonymous}}{}{%
  \ifthenelse{\equal{\paperversion}{camera-ready}}{\iclrfinalcopy}{%
    \ifthenelse{\equal{\paperversion}{arxiv}}{\iclrfinalcopy}{%
      \PackageError{WALA}{Invalid paperversion: \paperversion}
        {Use anonymous, camera-ready, or arxiv.}%
    }%
  }%
}
\ifthenelse{\equal{\paperversion}{arxiv}}{%
  \hypersetup{
    colorlinks=true,
    linkcolor=black,
    citecolor=black,
    urlcolor=blue,
    pdftitle={WALA: Learning Executable Latent Actions from Action-Labeled Demonstrations and Action-Free Videos},
    pdfauthor={Jiahao Liu, Yupeng Zheng, Zhongpu Xia, Shuai Tian, Huangrui Li, Yuhang Zheng, Ning Ma, ShangQing Zhou, Xiaotian Liu, Jing Li, Yixian Li, Haoran Li, Dongbin Zhao}
  }
}{}
\begin{document}
\ifthenelse{\equal{\paperversion}{arxiv}}{\WALApreprinttitle}{\maketitle}
\ifthenelse{\equal{\paperversion}{arxiv}}{\fancyhead[L]{Preprint}}{}

\begin{abstract}
Human videos provide rich information about object manipulation at a scale
difficult to reproduce with robots, but often lack action annotations that
can directly supervise robot policies.
Realizing this potential requires distinguishing interaction-relevant
changes from static appearance and connecting the learned interaction
knowledge to executable robot actions.
We present \textbf{WALA}, a framework named for
\emph{World- and Action-supervised Latent Actions}.
Its semantic-geometric latent action model (LAM) learns latent actions
by encoding and predicting semantic and geometric changes,
emphasizing interactions over static appearance while retaining
spatial detail.
During robot policy training, the LAM encoder and decoder provide
guidance through distillation and visual prediction, respectively.
Together with robot action supervision, these signals shape
policy-generated latent actions to support world prediction and
capture the information needed to directly generate executable robot actions.
Because LAM supervision requires no action labels,
WALA enables co-training on action-labeled robot demonstrations
and task-relevant action-free videos.
In simulation experiments, our method achieves \textbf{92.33\%}
average success on RoboTwin and \textbf{75.2\%} on
RoboCasa-GR1-Tabletop, exceeding the previous state of the art
on RoboCasa by \textbf{5.0} percentage points.
WALA also shows strong real-robot performance, further enhanced
by task-relevant action-free human videos that improve data
efficiency and support target-task zero-shot and few-shot transfer.
\end{abstract}

\begin{center}
\small\textbf{Project website:}\quad
\url{https://liujiahao2077.github.io/WALA.github.io}
\end{center}

\section{Introduction}

\suppressfloats[t]
\begin{figure}[t]
    \centering
    \includegraphics[width=\linewidth]{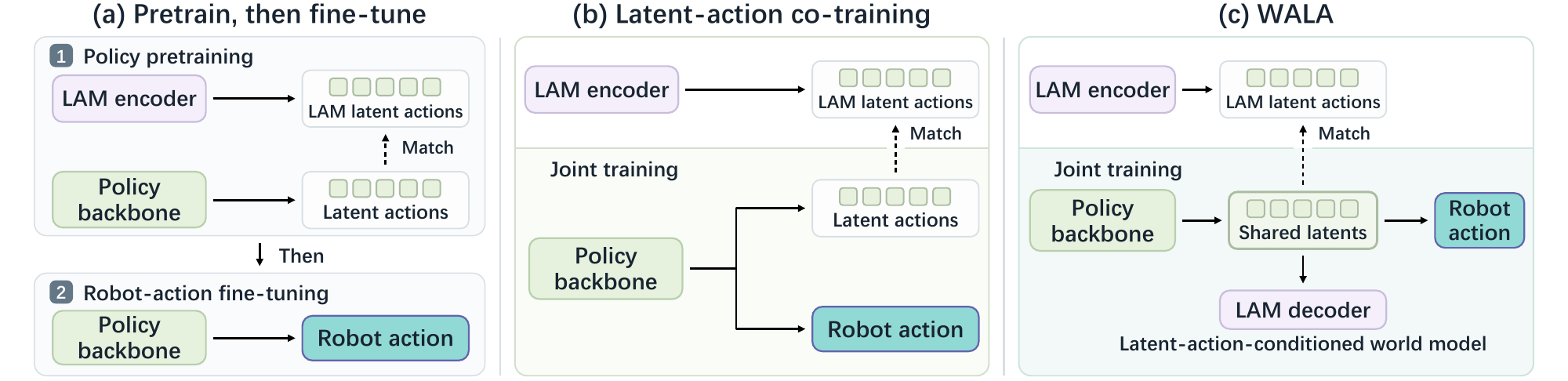}
    \caption{Three paradigms for robot policy learning with a pretrained LAM.
    (a) A policy is pretrained to predict latent actions and then fine-tuned
    on action-labeled robot demonstrations.
    (b) Co-training combines latent-action prediction and robot-action
    generation within the same policy.
    (c) WALA combines LAM encoder distillation, LAM decoder world prediction,
    and robot action supervision through shared policy latent actions
    that directly feed the action head.}
    \label{fig:wala_method_comparison}
\end{figure}

Generalist robot policies have shown strong performance on diverse manipulation
tasks~\citep{intelligence2025pi05visionlanguageactionmodelopenworld,
black2026pi0visionlanguageactionflowmodel}, but collecting the
robot demonstrations needed to extend these capabilities across tasks and
environments remains costly~\citep{khazatsky2024droid}. Egocentric human videos
can capture manipulation experience without operating a robot~\citep{kareer2025egomimic},
with public corpora spanning tens of thousands of hours and
beyond~\citep{grauman2022ego4d,grauman2024egoexo4d,
buildai2025egocentric100k,
hoque2025egodexlearningdexterousmanipulation}. Recorded or estimated human motion can provide
action supervision for policy pretraining or co-training with robot
demonstrations~\citep{luo2026beingh0,lyu2026lda1bscalinglatentdynamics}.
Since many videos contain no action
annotations~\citep{ye2025latentactionpretrainingvideos}, the complicated
pipeline which includes trajectory estimation, coordinate alignment, or
retargeting~\citep{yang2026egovla,singh2025handobject} is required.
These costs and cumulative errors raise the question of whether interaction
representations can instead be learned directly from unlabeled human videos,
without requiring explicit action annotations for robot policy learning.

Latent action models (LAMs) provide a way to learn interaction
representations from videos without action labels.
By modeling inverse and forward dynamics in observation space,
they learn actions in a latent space.
These models are commonly constructed directly in the raw pixel
space of visual
observations~\citep{ye2025latentactionpretrainingvideos,routray2026vipra}.
However, the redundant background information in raw observations
motivates the use of pretrained encoders to construct semantically
rich feature spaces, offering a more efficient alternative for
latent action learning~\citep{bu2025univlalearningacttaskcentric}.
Nevertheless, semantic information alone is insufficient to
accurately capture the spatial changes induced by actions, which
are crucial for robot manipulation~\citep{yang2024depthanything}.
Meanwhile, these LAMs focus on reconstructing complete future
observations in pixel or feature space, rather than specifically
capturing interaction-induced changes.
Our goal, however, is for LAMs to represent the actions that drive
these changes, rather than the states that result from them.

Once a LAM has learned meaningful interaction representations,
the next question is how to effectively transfer this knowledge
to robot control.
Such transfer can begin by pretraining a policy to predict
LAM-inferred latent actions, enabling it to acquire interaction
priors from videos without using ground-truth action labels.
The pretrained policy is then fine-tuned on action-labeled robot
demonstrations~\citep{ye2025latentactionpretrainingvideos}
(Fig.~\ref{fig:wala_method_comparison}(a)).
Rather than predicting latent actions only during policy
pretraining, co-training combines latent-action prediction
with robot action generation within the same
policy~\citep{bu2025univlalearningacttaskcentric,
chen2025motolatentmotiontoken},
allowing interaction knowledge to continue shaping the policy
(Fig.~\ref{fig:wala_method_comparison}(b)).
However, using the LAM only to provide encoder-derived latent
action targets does not exploit its decoder's potential to guide
downstream policy learning through visual prediction.
Beyond using the LAM to guide policy learning, we ask
whether latent actions themselves can provide the
information needed to directly generate executable robot actions.

These considerations highlight two challenges:
(1) learning latent actions that capture interaction-relevant changes
and spatial geometry, and
(2) effectively transferring this knowledge to robot action generation.
To address both, we introduce WALA, a framework named for
\emph{World- and Action-supervised Latent Actions}.
To address the first challenge, WALA's semantic-geometric LAM
encodes semantic-feature and depth-map deltas between current and
sparsely sampled future observations into latent actions,
and uses these latent actions to predict the same deltas.
Using temporal differences in both encoding and prediction
emphasizes interaction changes over shared static
appearance~\citep{wang2021tdn,chen2025temporal},
while pretrained visual features and depth-map supervision
provide complementary interaction semantics and fine-grained geometry.
To address the second challenge, WALA uses the pretrained LAM
together with robot action supervision to shape policy-generated
latent actions (Fig.~\ref{fig:wala_method_comparison}(c)).
The LAM encoder transfers pretrained interaction knowledge through
distillation, while its decoder serves as a latent-action-conditioned
world model that further refines these representations through
downstream interaction prediction.
The action head takes the same latent actions as its sole input,
grounding the interaction knowledge they encode in executable robot actions.
Since both forms of LAM supervision require no action labels,
WALA enables co-training on action-labeled robot demonstrations
and task-relevant action-free videos to further improve robot manipulation.

Our contributions are summarized as follows.

\begingroup
\setlength{\leftmargini}{1em}
\setlength{\labelsep}{0.45em}
\begin{itemize}
    \item We propose WALA, a framework that enables co-training on
    action-labeled demonstrations and action-free videos through a
    shared latent action representation.
    Its pretrained LAM guides this representation through encoder
    distillation and decoder prediction, while robot action supervision
    grounds it in executable robot actions.

    \item We design a semantic-geometric LAM that learns
    interaction-relevant latent actions by encoding and predicting
    semantic-feature and depth-map deltas.
    This formulation emphasizes interaction changes over shared static
    appearance, combining semantic abstraction with depth-map
    supervision to preserve fine-grained geometry.

    \item We demonstrate strong performance across simulation and
    real-world manipulation. Our method achieves 92.33\% average
    success on RoboTwin and 75.2\% on RoboCasa-GR1-Tabletop.
    Real-robot experiments further show that task-relevant action-free
    egocentric human videos improve performance and data efficiency
    while supporting target-task zero-shot and few-shot transfer.
\end{itemize}
\endgroup

\section{Related Work}

\subsection{Vision-Language-Action Models and World-Action Models}

Vision-language-action (VLA) models such as
OpenVLA~\citep{kim2024openvlaopensourcevisionlanguageactionmodel},
$\pi_0$~\citep{black2026pi0visionlanguageactionflowmodel}, and
$\pi_{0.5}$~\citep{intelligence2025pi05visionlanguageactionmodelopenworld}
map visual observations and language instructions to actions across tasks and
embodiments. QDepth-VLA~\citep{li2026qdepthvla} augments VLA training with
auxiliary quantized-depth prediction for geometric perception.
World-action models (WAMs) jointly predict actions and future observations.
LingBot-VA~\citep{li2026causalworldmodelingrobot},
DreamZero~\citep{ye2026worldactionmodelszeroshot},
Motus~\citep{Bi_2026_CVPR}, and
Fast-WAM~\citep{yuan2026fastwamworldactionmodels} predict future RGB observations,
typically through compressed video latents, retaining background appearance,
texture, and illumination in the prediction targets.
LDA-1B~\citep{lyu2026lda1bscalinglatentdynamics} instead predicts dynamics in
a structured DINO feature space.
WALA combines robot action learning with world-model supervision,
predicting future DINO-feature and depth-map deltas to capture
interaction semantics and local geometry.

\subsection{Latent Action Learning from Videos}

Latent action models differ in their prediction targets and representations.
LAPA~\citep{ye2025latentactionpretrainingvideos} and
Moto~\citep{chen2025motolatentmotiontoken} learn discrete visual-motion codes
through future-frame reconstruction. ViPRA~\citep{routray2026vipra} augments
RGB reconstruction with perceptual and optical-flow consistency losses.
UniVLA~\citep{bu2025univlalearningacttaskcentric}
uses language-conditioned DINO-feature prediction.
villa-X~\citep{chen2025villaxenhancinglatentaction} couples visual reconstruction
with proprioceptive forward dynamics, while JALA~\citep{luo2026jointalignedlatentaction}
aligns policy predictions with inverse-dynamics latents and available actions.
UniT~\citep{chen2026unitunifiedphysicallanguage} combines visual, action, and
fused branches through shared quantization and cross-reconstruction.
WALA learns latent actions from complementary semantic-feature
and depth-map deltas between the current and multiple
sparsely sampled future frames.

The learned latent actions are also incorporated into robot policy learning in
different ways.
LAPA~\citep{ye2025latentactionpretrainingvideos} directly fine-tunes on robot actions, whereas
UniVLA~\citep{bu2025univlalearningacttaskcentric} retains latent-action prediction
with embodiment-specific action decoding.
Moto~\citep{chen2025motolatentmotiontoken} jointly fine-tunes latent-motion and
robot-action prediction.
UniT~\citep{chen2026unitunifiedphysicallanguage} provides latent action token targets for
VLA learning and action conditions for world modeling.
ViPRA~\citep{routray2026vipra} jointly pretrains visual-token and latent-action
prediction before learning robot actions through flow matching.
WALA applies encoder distillation and semantic-geometric decoder prediction
to policy-generated latent action tokens that directly feed the action head.

\raggedbottom
\section{Method}

\begin{figure}[!t]
    \centering
    \includegraphics[width=\linewidth]{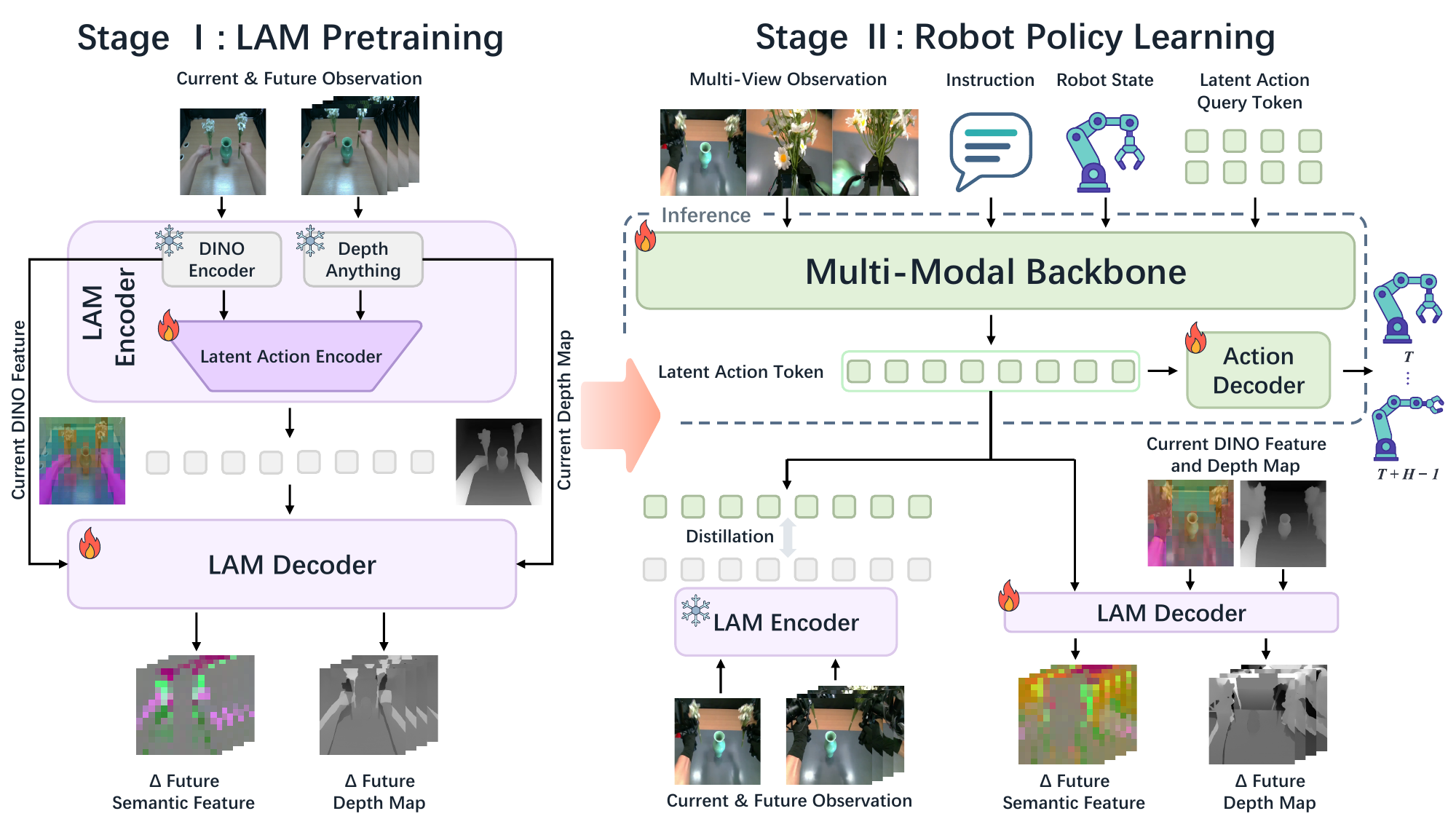}
    \caption{
    Overview of WALA. Stage I pretrains the semantic-geometric LAM to encode
    and predict changes between current and sparsely sampled future
    primary-view observations. Stage II jointly shapes policy-generated
    latent actions through LAM encoder distillation, LAM decoder prediction,
    and robot action supervision. These tokens form the action head's sole
    input. The dashed box marks the inference pathway.
    Future observations and the LAM are used only during training.
    Snowflakes and flames denote frozen and trainable modules, respectively.
    }
    \label{fig:wala_overview}
\end{figure}

\subsection{Overview}

Our goal is to learn a policy from action-labeled robot demonstrations and
task-relevant action-free videos, mapping current observations and
instructions to executable actions. WALA connects these data sources
through shared latent actions in two stages (Fig.~\ref{fig:wala_overview}).
Stage I pretrains a semantic-geometric LAM to encode and predict visual
changes without action labels. Stage II co-trains the policy on both data
sources, shaping policy-generated latent actions through LAM encoder
distillation, LAM decoder prediction, and robot action supervision.

The policy receives multi-view RGB observations, while the LAM uses only the
primary camera view. Let $I_t$ denote its current RGB image and $D_t$ its
normalized depth map from frozen Depth Anything V2~\citep{yang2024depthv2}
(Appendix~\ref{sec:loss_implementation}). We sample
$K=4$ future observations $\{o_{t+\tau_k}\}_{k=1}^{K}$, where
$o_{t+\tau_k}=(I_{t+\tau_k},D_{t+\tau_k})$, at evenly spaced offsets within each clip.
During robot policy training, human and robot clips cover the same temporal
extent as an action chunk (Appendix~\ref{sec:video_policy_inputs}). A frozen DINOv3~\citep{simeoni2025dinov3}
encoder $\phi$ extracts patch-level semantic features,
\begin{equation}
    X_t = \phi(I_t), \quad
    X_{t,k}^{+} = \phi(I_{t+\tau_k}) .
\end{equation}
We write $X_t^{+}=\{X_{t,k}^{+}\}_{k=1}^{K}$ and
$D_t^{+}=\{D_{t,k}^{+}\}_{k=1}^{K}$ for the future features and depth maps,
with $D_{t,k}^{+}=D_{t+\tau_k}$. The corresponding deltas are
$\Delta X_{t,k}=X_{t,k}^{+}-X_t$ and
$\Delta D_{t,k}=D_{t,k}^{+}-D_t$, collected in
$\Delta X_t^{+}$ and $\Delta D_t^{+}$. Throughout, $+$ denotes future
observations, whereas a hat denotes a model prediction.

\subsection{Stage I: LAM Pretraining}

The semantic-geometric LAM learns interaction-relevant latent actions by
encoding and predicting temporal deltas. Its encoder $E_{\theta}$ processes
the current observation and future deltas through semantic and geometric
branches, then fuses them into latent action tokens,
\begin{equation}
    z_t = E_{\theta}(X_t, D_t, \Delta X_t^{+}, \Delta D_t^{+}) .
\end{equation}
The semantic branch operates in patch-level DINOv3 feature space, while the
geometric branch operates directly in depth-map space.
The representation $z_t$ comprises 32 tokens of dimension 1,024, compressing
the changes across all four future observations.
These tokens summarize visual changes over the sampled clip; they are not
per-step robot commands. Their mapping to executable control is learned during
policy training.

The LAM decoder $G_{\psi}$ predicts future deltas
conditioned on both the current observation and $z_t$,
\begin{equation}
\label{eq:lam_decode}
    \widehat{\Delta X}_{t,k}
    =
    G^{\mathrm{sem}}_{\psi}(X_t, z_t, k),
    \quad
    \widehat{\Delta D}_{t,k}
    =
    G^{\mathrm{dep}}_{\psi}(D_t, z_t, k).
\end{equation}
Here, $\theta$ and $\psi$ parameterize the encoder and decoder;
$k\in\{1,\ldots,K\}$ indexes the future observation at $t+\tau_k$.
Both branches use latent-action-conditioned feature-wise linear
modulation (FiLM)~\citep{perez2018film} to modulate their respective current
features. By reusing the current observation, this design
reduces the need to encode shared scene content in the latent actions
and encourages them to represent changes relative to the current state.
Adding the predicted deltas to the current semantic features and depth maps
yields their future counterparts. DINOv3 feature prediction provides semantic
supervision without future RGB reconstruction, while direct depth-map
prediction supplies spatially detailed geometric supervision.

For brevity, $\Delta X$ and $\Delta D$ below collect all sampled future
deltas, with time and future-set indices omitted.
The semantic objective combines element-wise regression with cosine loss,
which additionally constrains the direction of feature change,
\begin{equation}
    \mathcal{L}_{\mathrm{sem}}
    =
    \mathcal{L}_{\mathrm{reg}}^{\mathrm{sem}}
    (\widehat{\Delta X}, \Delta X)
    +
    \lambda_{\mathrm{cos}}
    \mathcal{L}_{\mathrm{cos}} .
\end{equation}
The depth objective combines regression with depth-gradient consistency,
\begin{equation}
    \mathcal{L}_{\mathrm{dep}}
    =
    \mathcal{L}_{\mathrm{reg}}^{\mathrm{dep}}
    (\widehat{\Delta D}, \Delta D)
    +
    \lambda_{\mathrm{grad}}
    \mathcal{L}_{\mathrm{grad}} .
\end{equation}
The gradient term matches horizontal and vertical finite differences of the
predicted and target depth deltas to supervise local depth variations. Losses are
averaged over future steps and spatial locations. The coefficients
$\lambda_{\mathrm{cos}}$ and $\lambda_{\mathrm{grad}}$ weight these auxiliary terms.
The LAM pretraining objective is
\begin{equation}
\label{eq:lam_objective}
    \mathcal{L}_{\mathrm{LAM}}
    =
    \mathcal{L}_{\mathrm{sem}}
    +
    \lambda_{\mathrm{dep}}
    \mathcal{L}_{\mathrm{dep}} .
\end{equation}
Here, $\lambda_{\mathrm{dep}}$ weights depth relative to semantic prediction.
Exact loss definitions and all weights are given in
Appendix~\ref{sec:loss_implementation}.

\subsection{Stage II: Robot Policy Learning}
\label{sec:policy_training}

After pretraining, we freeze the LAM encoder to provide a stable latent action
target space. It encodes each sample's observed changes as a distillation target,
\begin{equation}
    z_t^{\mathrm{ref}}
    =
    \mathrm{sg}
    \left[
    E_{\theta}(X_t, D_t, \Delta X_t^{+}, \Delta D_t^{+})
    \right],
\end{equation}
where $\mathrm{sg}[\cdot]$ denotes stop-gradient.

The policy combines a multimodal backbone with an action-query interface.
Backbone variants are described in Appendix~\ref{sec:policy_backbones}.
Given multi-view RGB observations $o_t^{\mathrm{mv}}$, a language
instruction $l$, robot state $s_t$ when available, and latent action queries
$q$, the policy backbone $\Pi_{\omega}$ produces $H$ latent action tokens,
one per step of an action chunk. The action head $A_{\omega}$ takes only
these tokens as input,
\begin{equation}
    \widehat{a}_{t:t+H-1} = A_{\omega}(\widetilde{z}_t),
    \quad
    \widetilde{z}_t = \Pi_{\omega}(o_t^{\mathrm{mv}}, l, s_t, q).
\end{equation}
Here, $\omega$ collects backbone and action-head parameters, and a tilde
identifies policy-generated latent actions. Two learned adapters align the token count and feature dimension
of $\widetilde z_t$ with those of the LAM latent actions for
encoder-target distillation and decoder conditioning
(Appendix~\ref{sec:model_training_configuration}):
\begin{equation}
    \bar{z}^{\mathrm{enc}}_t=P_{\mathrm{enc}}(\widetilde{z}_t),
    \qquad
    \bar{z}^{\mathrm{dec}}_t=P_{\mathrm{dec}}(\widetilde{z}_t).
\end{equation}
Writing $U=\bar z_t^{\mathrm{enc}}$ and $V=z_t^{\mathrm{ref}}$, the distillation loss is
\begin{equation}
\label{eq:distillation}
\mathcal{L}_{\mathrm{distill}}
=\ell_{\mathrm{cos}}(U,V)
+0.1\operatorname{SmoothL1}\!\left(\operatorname{LN}(U),\operatorname{LN}(V)\right),
\end{equation}
where $\ell_{\mathrm{cos}}$ averages token-wise cosine distance and
$\operatorname{LN}$ denotes layer normalization over each token's features
(Appendix~\ref{sec:loss_implementation}).
The pretrained LAM decoder serves as a latent-action-conditioned
semantic-geometric world model. It receives $\bar z_t^{\mathrm{dec}}$ in
place of $z_t$ in Eq.~(\ref{eq:lam_decode}) and is jointly fine-tuned with
the policy. Its loss $\mathcal{L}_{\mathrm{wm}}$ has the same form and
internal weights as $\mathcal{L}_{\mathrm{LAM}}$ in
Eq.~(\ref{eq:lam_objective}), evaluated on predictions conditioned on
policy-generated rather than LAM-encoded latent actions.
Distillation transfers pretrained interaction knowledge, while decoder
prediction supervises the interaction outcomes implied by policy-generated
latent actions. Both losses refine $\widetilde z_t$ through the adapters.
With the action head reading only these tokens, world and robot action
supervision jointly shape the representation used for interaction
prediction and executable action generation.

At the sample level, the available supervision is expressed as
\begin{equation}
    \mathcal{L}_{\mathrm{policy}}
    =
    m\mathcal{L}_{\mathrm{act}}
    +
    \lambda_{\mathrm{distill}}\mathcal{L}_{\mathrm{distill}}
    +
    \lambda_{\mathrm{wm}}\mathcal{L}_{\mathrm{wm}} .
\end{equation}
The indicator $m$ is 1 for samples with robot action labels and 0 otherwise.
The action loss $\mathcal{L}_{\mathrm{act}}$ is mean L1 error on action chunks.
In minibatch optimization, action losses are averaged over labeled samples,
whereas both LAM losses are averaged over all samples. The latter support
co-training on action-labeled demonstrations and task-relevant action-free
videos, jointly refining the latent actions used for robot control.
Future observations supervise training but never enter the policy backbone.
The coefficients $\lambda_{\mathrm{distill}}$ and $\lambda_{\mathrm{wm}}$
weight the auxiliary losses; their selection and the minibatch reductions
are detailed in Appendix~\ref{sec:loss_implementation}.

\subsection{Training and Inference}

Stage I optimizes the LAM encoder, fusion module, and decoder while keeping
DINOv3 and the depth estimator frozen. Stage II keeps the LAM encoder
frozen and jointly optimizes the decoder, adapters, policy backbone, and action
head, adapting the world model to policy tokens and downstream observations.
At inference, only the policy backbone and action head are used to map current
multi-view RGB observations, language, available robot state, and action
queries to robot actions.

\section{Experiments}

Our experiments address four questions:
(1) How does WALA compare with VLA-, WAM-, and LAM-based policies?
(2) How does performance scale with action-labeled demonstrations and
action-free videos?
(3) What interaction and action information do the latent actions retain,
and how do their supervision and action readout contribute to robot control?
(4) Can WALA achieve effective and efficient real-robot manipulation and
use task-relevant action-free videos to improve performance and transfer?

\subsection{Experimental Setup}
\label{sec:experimental_setup}

Standard WALA uses Qwen3-VL-4B-Instruct~\citep{bai2025qwen3vltechnicalreport};
WALA-Wan uses Wan2.2-TI2V-5B~\citep{wanai2025wan22ti2v5b}.
RoboTwin~\citep{chen2025robotwin20scalabledata} uses 50 Clean and 500 Random
demonstrations per task across 50 tasks, with 100 evaluation episodes per
task in each setting. RoboCasa-GR1-Tabletop~\citep{nasiriany2024robocasalargescalesimulationeveryday}
uses 1{,}000 demonstrations and 50 evaluation episodes for each of 24 tasks.
Our action-only base policy shares WALA's backbone and robot demonstrations,
isolating the contribution of LAM supervision.

\begin{figure}[!b]
    \centering
    \includegraphics[width=0.49\linewidth]{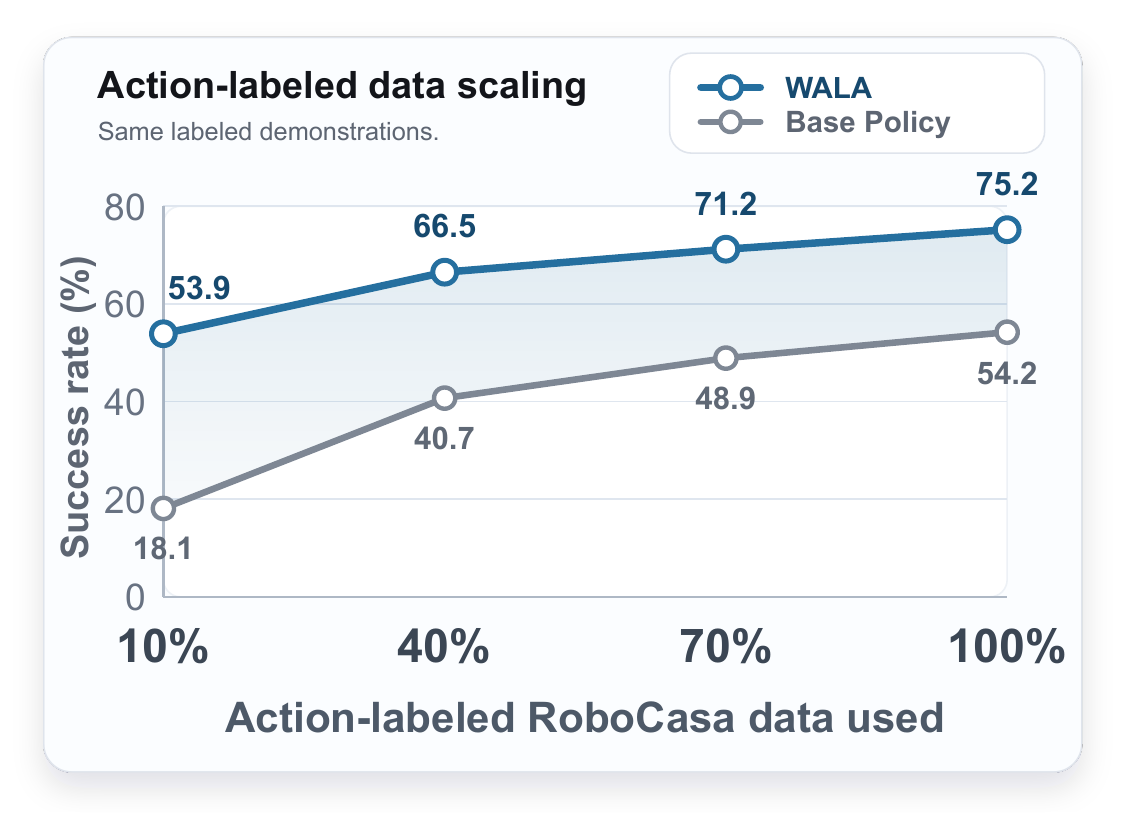}%
    \hfill%
    \includegraphics[width=0.49\linewidth]{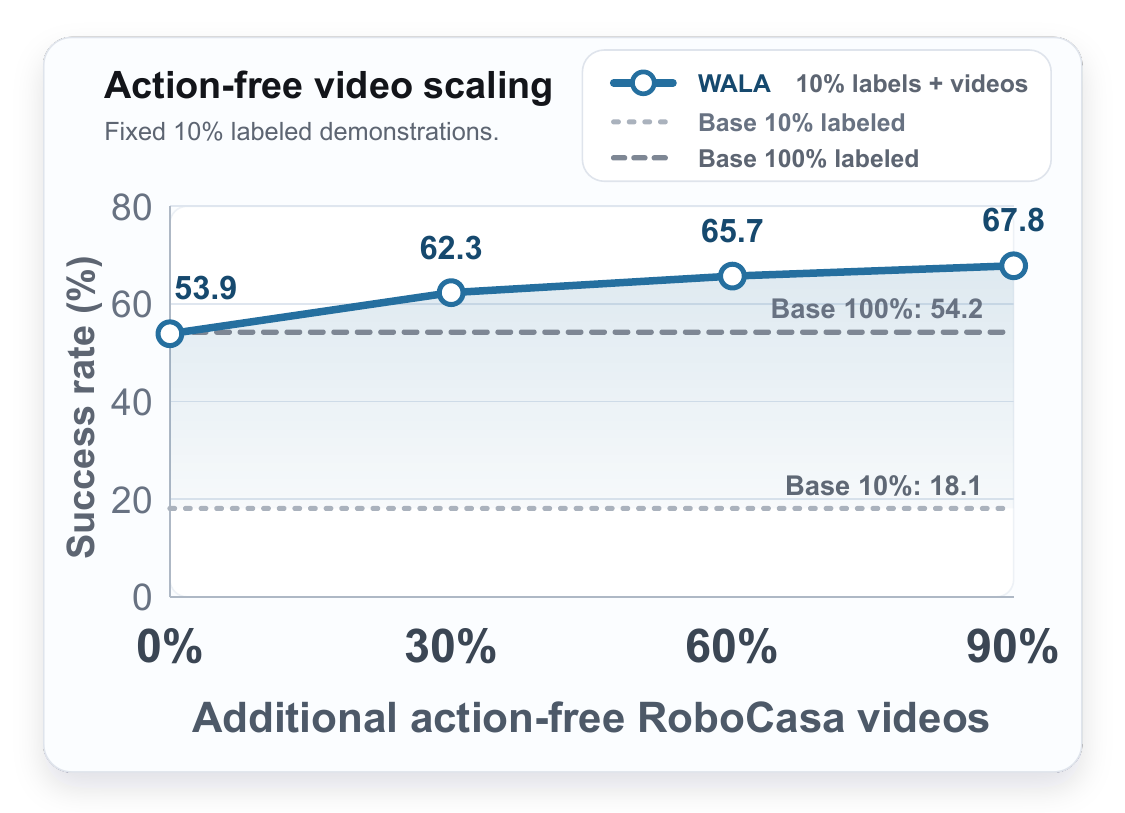}
    \caption{
    Data scaling on RoboCasa-GR1-Tabletop. Left: WALA consistently improves
    over the base policy when both use the same number of action-labeled
    demonstrations. Right: with 10\% action-labeled demonstrations fixed,
    additional videos used without action supervision further improve WALA.
    All data fractions refer to the full RoboCasa training set.
    }
    \label{fig:data_scaling}
\end{figure}

\newpage
\subsection{Main Benchmark Results}

\begin{wraptable}{l}{.48\textwidth}
\centering
\caption{
Main results on RoboTwin. Baseline results are collected from
publicly released reports. Avg. is the mean of the Clean and Random success
rates.
}
\label{tab:robotwin_main}
\scriptsize
\renewcommand{\arraystretch}{1.08}
\setlength{\tabcolsep}{3.2pt}
\resizebox{\linewidth}{!}{%
\begin{tabular}{llccc}
\toprule
Group & Method & Clean & Random & Avg. \\
\midrule
\multirow{4}{*}{\makecell[l]{VLA-based}}
& $\pi_{0.5}$~\citep{intelligence2025pi05visionlanguageactionmodelopenworld}
                   & 82.74 & 76.76 & 79.75 \\
& X-VLA~\citep{zheng2025xvlasoftpromptedtransformerscalable}
                   & 72.80 & 72.84 & 72.82 \\
& StarVLA-$\alpha$~\citep{ye2026starvlaalphareducingcomplexityvisionlanguageaction}
                   & 88.20 & 88.30 & 88.25 \\
& InternVLA-A1~\citep{cai2026internvlaa1unifyingunderstandinggeneration}
                   & 89.40 & 89.64 & 89.52 \\
\midrule
\multirow{3}{*}{\makecell[l]{WAM-based}}
& Motus~\citep{Bi_2026_CVPR}
                   & 88.66 & 87.02 & 87.84 \\
& Fast-WAM~\citep{yuan2026fastwamworldactionmodels}
                   & 91.88 & 91.78 & 91.83 \\
& LingBot-VA~\citep{li2026causalworldmodelingrobot}
                   & \best{92.93} & 91.55 & \second{92.24} \\
\midrule
\rowcolor{WALABlue}
     & WALA & \walaRobotwinClean
           & \best{\walaRobotwinRandom}
           & \walaRobotwinAvg \\
\rowcolor{WALABlue}
\multirow{-2}{*}{Ours}
     & WALA-Wan & \second{\walaWanRobotwinClean}
           & \second{\walaWanRobotwinRandom}
           & \best{\walaWanRobotwinAvg} \\
\bottomrule
\end{tabular}
}
\par\vspace{10pt}
\centering
\caption{
Main results on RoboCasa-GR1-Tabletop.
$^{\dagger}$ marks our reproductions; other baselines are from public reports.
}
\label{tab:robocasa_main}
\small
\renewcommand{\arraystretch}{1.08}
\setlength{\tabcolsep}{7pt}
\resizebox{\linewidth}{!}{%
\begin{tabular}{llc}
\toprule
Group & Method & Avg. \\
\midrule
\multirow{8}{*}{\makecell[l]{VLA-based}}
& StarVLA Qwen3-PI~\citep{community2026starvlalegolikecodebasevisionlanguageaction}
                         & 43.9 \\
& GR00T-N1.6~\citep{nvidia2025gr00tn1openfoundation}
                         & 47.6 \\
& StarVLA Qwen3-GR00T~\citep{community2026starvlalegolikecodebasevisionlanguageaction}
                         & 47.8 \\
& StarVLA Qwen3-OFT~\citep{community2026starvlalegolikecodebasevisionlanguageaction}
                         & 48.8 \\
& StarVLA-$\alpha$~\citep{ye2026starvlaalphareducingcomplexityvisionlanguageaction}
                         & 57.3 \\
& ABoT-M0~\citep{yang2026abotm0vlafoundationmodel}
                         & 58.3 \\
& RLDX-1~\citep{kim2026rldx1technicalreport}
                         & 58.7 \\
& FrameSkip~\citep{yu2026frameskiplearningfewerinformative}
                         & 59.5 \\
\midrule
\multirow{3}{*}{\makecell[l]{WAM-based}}
& DiT4DiT~\citep{ma2026dit4ditjointlymodelingvideo}
                         & 50.8 \\
& LDA-1B~\citep{lyu2026lda1bscalinglatentdynamics}
                         & 55.4 \\
& DIAL~\citep{chen2026dialdecouplingintentaction}
                         & \second{70.2} \\
\midrule
\multirow{3}{*}{\makecell[l]{LAM-based}}
& Moto~\citep{chen2025motolatentmotiontoken}$^{\dagger}$
                         & 45.6 \\
& UniVLA~\citep{bu2025univlalearningacttaskcentric}$^{\dagger}$
                         & 51.2 \\
& VLA-UniT~\citep{chen2026unitunifiedphysicallanguage}
                         & 66.7 \\
\midrule
\rowcolor{WALABlue}
Ours
& WALA                  & \best{75.2} \\
\bottomrule
\end{tabular}
}
\end{wraptable}

\textbf{RoboTwin.}
WALA-Wan leads all VLA- and WAM-based baselines with 92.33\% average success
(Table~\ref{tab:robotwin_main}); both variants exceed 92\% in Random.
LAM encoder distillation transfers pretrained interaction knowledge, while
LAM decoder prediction refines it using downstream observations. Joint
robot action supervision grounds this interaction-informed representation
in the actions generated by the policy. Semantic-geometric delta modeling
reduces emphasis on shared background appearance, helping maintain
performance under changes in backgrounds, lighting, and distractors.
The strong results with both backbones support this training framework
across vision-language and video policies.

\noindent\textbf{RoboCasa-GR1-Tabletop.}
WALA reaches 75.2\%, leading all three baseline groups
(Table~\ref{tab:robocasa_main}) and surpassing the previous
state-of-the-art method, DIAL, by 5.0 percentage points.
Among LAM-based methods, WALA exceeds VLA-UniT by 8.5 points and
outperforms our Moto and UniVLA reproductions under the same GR1 data
and evaluation protocol. Unlike RoboTwin, where strong baselines already
exceed 90\%, RoboCasa leaves more room for improvement. These tasks require
identifying the instructed object and receptacle, then placing the object
correctly. Semantic supervision supports target identification, while depth
supervision represents object--receptacle geometry. Rich human hand--object
interactions in pretraining also offer transferable manipulation knowledge
for GR1's dexterous hands. WALA also exceeds LDA-1B (55.4\%),
which directly pretrains robot policies on large-scale human videos with
aligned action annotations, demonstrating effective use of human experience
without human action labels.

\subsection{Scaling with Robot Demonstrations and Action-Free Videos}

\noindent\textbf{Robot-demonstration scaling.}
With 10\% action-labeled demonstrations, WALA achieves 53.9\% success,
compared with 18.1\% for the base policy at the same budget, nearly matching
the base policy's 54.2\% with all demonstrations
(Fig.~\ref{fig:data_scaling}, left). With limited robot demonstrations,
action supervision alone may be insufficient to learn rich interaction
representations. Knowledge learned from videos and supervision of semantic
and geometric changes provide additional information about object motion
and spatial relationships, helping the policy make better use of the
available robot demonstrations. As more action-labeled demonstrations
become available, both policies improve and the base policy narrows the
gap, consistent with robot data providing increasingly broad coverage
of these interactions. Nevertheless, WALA retains a 21.0-point advantage
at the full budget, showing that visual interaction supervision remains
beneficial even with substantially more robot action data.

\noindent\textbf{Video scaling without action supervision.}
With action-labeled demonstrations fixed at 10\%, success increases
steadily as more videos are added (Fig.~\ref{fig:data_scaling}, right).
These videos broaden the object configurations and interaction stages
encountered during policy training, extending visual supervision beyond
the small action-labeled subset. This richer experience helps the policy
learn task-relevant interaction patterns across different manipulation
examples, while the robot action supervision budget remains unchanged.
The resulting 13.9-point gain demonstrates the value of co-training on
action-labeled demonstrations and action-free videos. Both panels show diminishing returns,
consistent with increasing overlap among examples of the same tasks,
but videos continue to provide useful supervision at every increment.

\begin{table}[!t]
\centering
\caption{
Ablation study on RoboCasa-GR1-Tabletop.
}
\label{tab:ablation}
\small
\renewcommand{\arraystretch}{1.05}
\setlength{\tabcolsep}{3.6pt}
\begin{tabular}{lcccccc}
\toprule
\multirow{2}{*}{Variant}
& \multicolumn{4}{c}{Supervision}
& \multirow{2}{*}{Avg.}
& \multirow{2}{*}{Gain} \\
\cmidrule(lr){2-5}
& \makecell{LAM\\Pretraining}
& \makecell{Latent\\Distill.}
& \makecell{Semantic\\Feature Pred.}
& \makecell{Depth-Map\\Pred.}
& & \\
\midrule
Base Policy
& \xmark & \xmark & \xmark & \xmark & 54.2 & -- \\
+ World Model without Pretraining
& \xmark & \xmark & \cmark & \cmark & 67.8 & +13.6 \\
+ Semantic-Feature Prediction
& \cmark & \xmark & \cmark & \xmark & 68.8 & +14.6 \\
+ Semantic-Geometric World Model
& \cmark & \xmark & \cmark & \cmark & 71.0 & +16.8 \\
+ Latent-Action Distillation
& \cmark & \cmark & \xmark & \xmark & 67.6 & +13.4 \\
\midrule
\rowcolor{WALABlue}
WALA
& \cmark & \cmark & \cmark & \cmark
& \best{75.2}
& \best{+21.0} \\
\bottomrule
\end{tabular}
\end{table}

\subsection{Ablation Study}
\label{sec:ablation}

Table~\ref{tab:ablation} uses our action-only base policy (54.2\%) with the
same policy backbone and robot data. Distillation uses
$\mathcal{L}_{\mathrm{distill}}$; semantic and geometric prediction form
$\mathcal{L}_{\mathrm{wm}}$ (Sec.~\ref{sec:policy_training}). All rows retain
action supervision; variant configurations are detailed in
Appendix~\ref{sec:model_training_configuration}.
World-model supervision reaches 67.8\% without LAM
pretraining and 71.0\% with pretraining, showing that downstream prediction
benefits from prior interaction knowledge. The semantic-geometric variant
outperforms the variant using a separately pretrained semantic-only LAM by 2.2 points,
supporting geometry-aware latent action learning for spatial manipulation.
WALA reaches 75.2\%, exceeding distillation alone by 7.6 points and
world-model prediction alone by 4.2 points. Encoder targets transfer
pretrained knowledge, while decoder supervision refines the action-generating
tokens through prediction of downstream interactions. Their combined gain
shows that encoder-derived targets and downstream visual prediction provide
complementary guidance for the same action-generating representation.
On RoboDojo videos unseen during LAM pretraining,
decoder supervision also reduces semantic and depth prediction errors
relative to distillation alone; fine-tuning the decoder improves them further
(Appendix~\ref{sec:decoder_adaptation}). We further examine the action-readout design
(Appendix~\ref{sec:all_tokens_readout}). Reading all final-layer VLM
tokens through a learned resampler yields 73.3\% success, compared
with 75.2\% for WALA, providing no additional benefit from the broader
feature access. Latent action tokens also receive the highest attention in this variant, supporting their role as a
compact multimodal representation for direct robot action generation.

\subsection{Analysis of Learned Latent Actions}

\begin{figure}[!hbp]
    \centering
    \begin{minipage}[t]{0.32\textwidth}
        \centering
        {\normalsize\bfseries\strut (a) WALA\par}
        \vspace{2pt}\nointerlineskip
        \includegraphics[width=\linewidth]{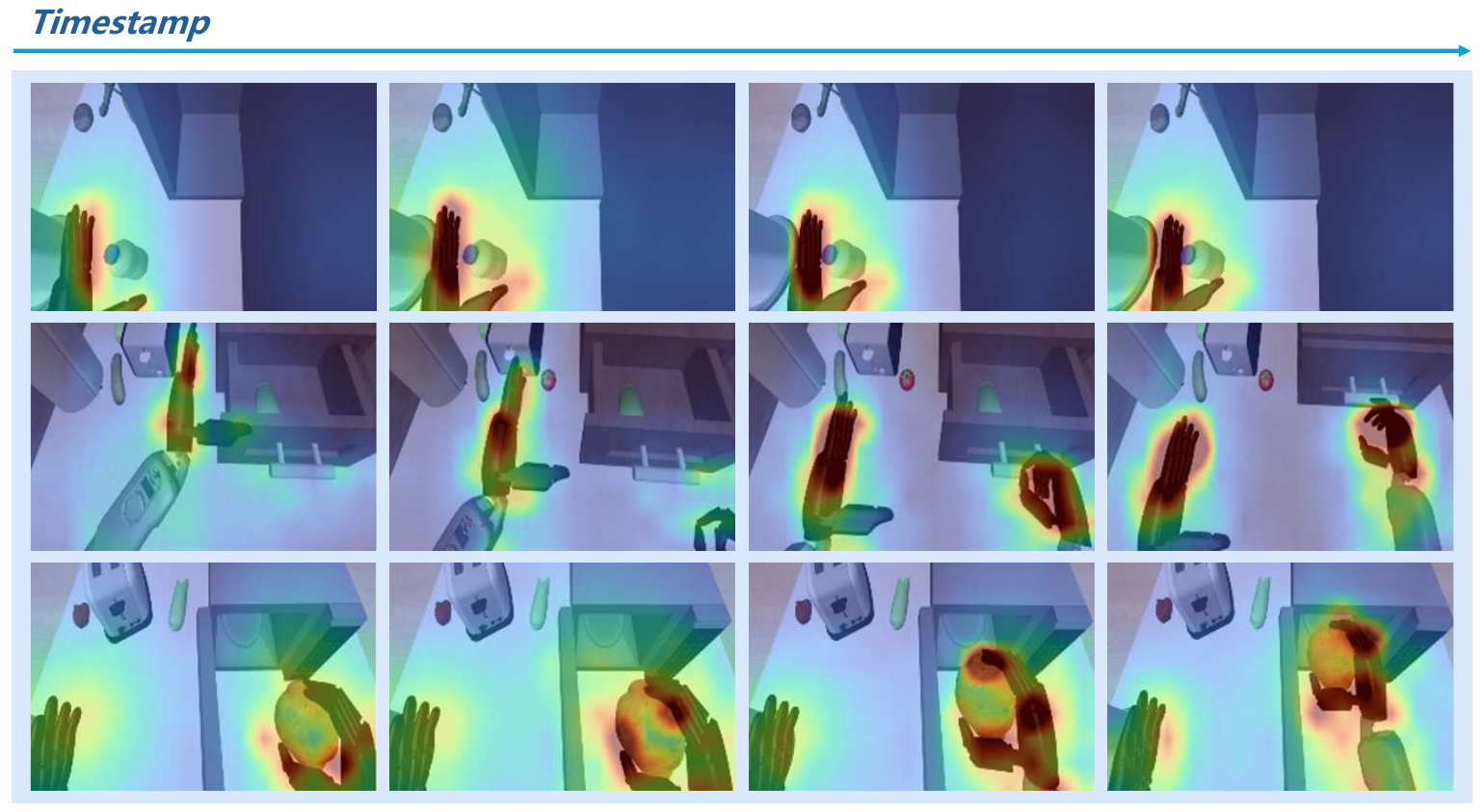}
    \end{minipage}\hfill
    \begin{minipage}[t]{0.32\textwidth}
        \centering
        {\normalsize\bfseries\strut (b) UniVLA\par}
        \vspace{2pt}\nointerlineskip
        \includegraphics[width=\linewidth]{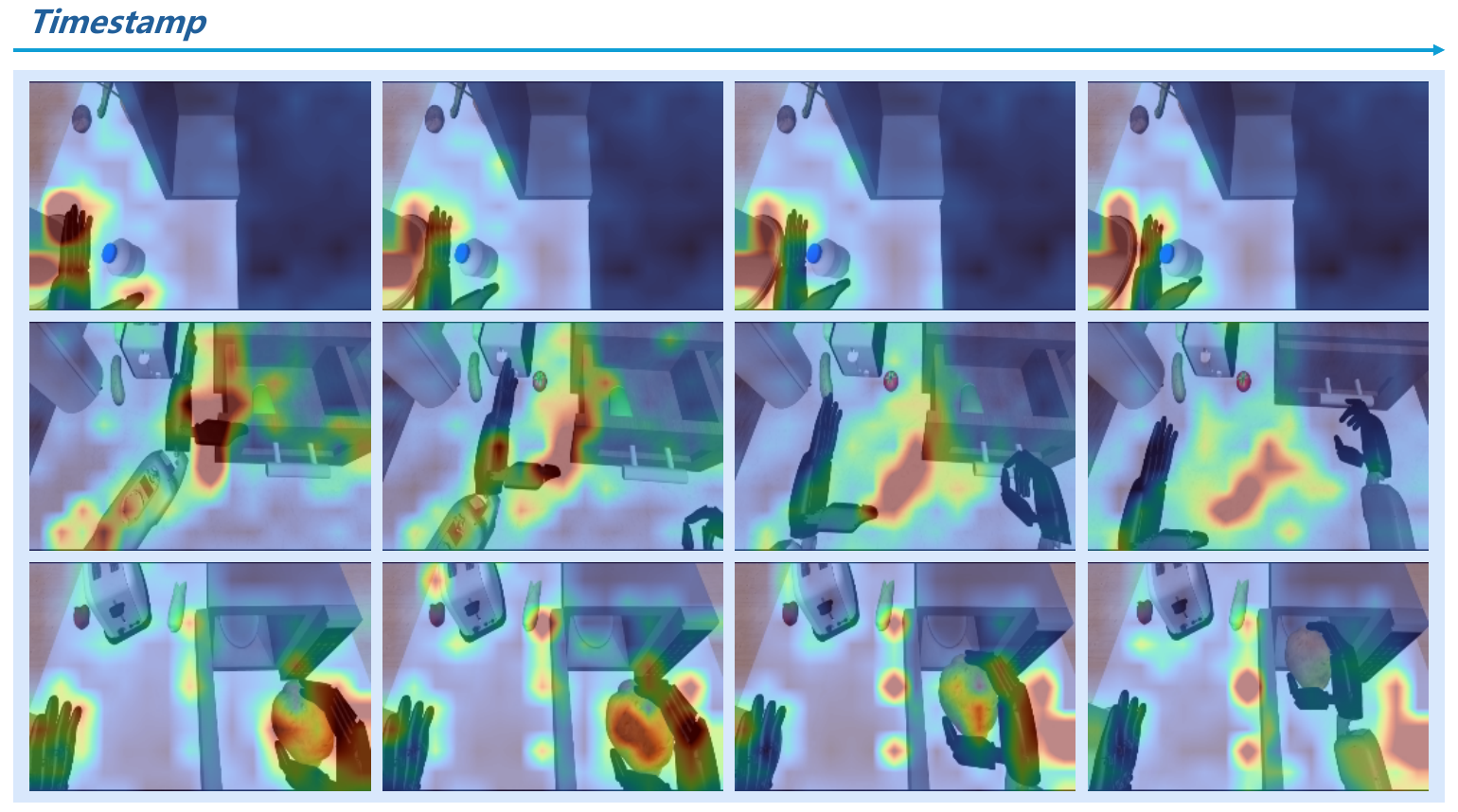}
    \end{minipage}\hfill
    \begin{minipage}[t]{0.32\textwidth}
        \centering
        {\normalsize\bfseries\strut (c) Moto\par}
        \vspace{2pt}\nointerlineskip
        \includegraphics[width=\linewidth]{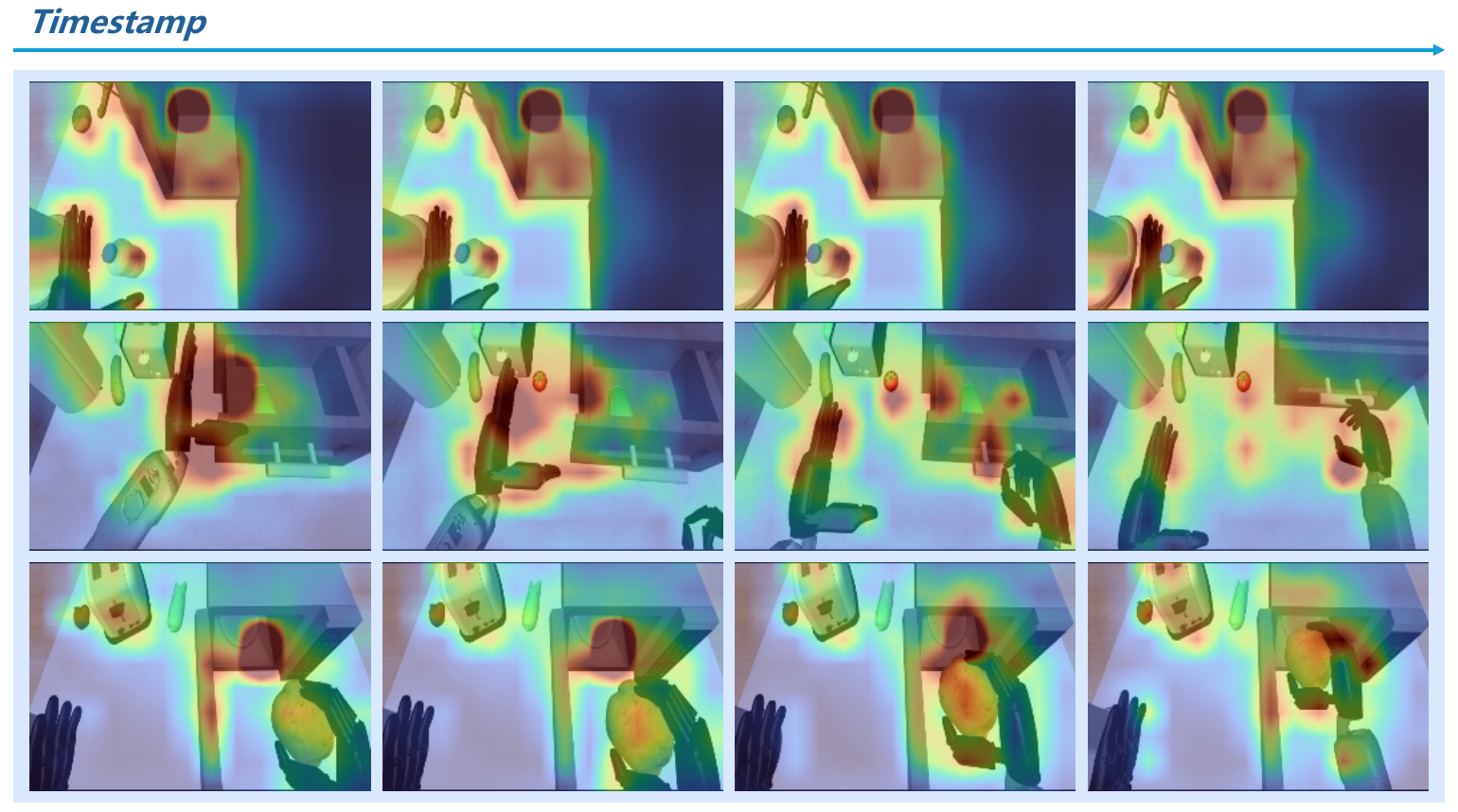}
    \end{minipage}
    \caption{Latent-action encoder attention from WALA, UniVLA, and Moto.
    Corresponding rows show the same four future frames of one RoboCasa task.}
    \label{fig:attention_vis}
\end{figure}

We first examine the policy-generated latent actions used for robot control.
In Appendix Fig.~\ref{fig:pred_vis}, the decoder's semantic and depth predictions
capture hand and object movements together with changing spatial structure,
showing that the action-generating tokens retain both forms of information.
With the current observation fixed, swapping tokens produces incorrect
semantic and geometric changes; swapping or zeroing them increases prediction errors
(Appendix~\ref{sec:token_dependency}). Thus, the tokens retain information
about future interactions that the decoder cannot obtain from the current
scene alone.

We next examine the interaction representations learned by the LAM encoder.
WALA's attention
concentrates on hands, manipulated objects, and their interaction regions,
whereas UniVLA's and Moto's are more dispersed across the scene
(Fig.~\ref{fig:attention_vis}). This interaction focus is consistent with
delta modeling and helps explain WALA's performance amid visual distractors.
Latent-action retrieval recovers similar
manipulation primitives across objects, layouts, and backgrounds
(Appendix~\ref{sec:latent_retrieval}). Action probes of the frozen LAM show that WALA latent actions support
more accurate robot action decoding than UniVLA and Moto latent actions
and purely visual feature sequences (Appendix~\ref{sec:action_probes}).
These results indicate that WALA's latent actions retain not only
world interaction information but also information that can be
effectively mapped to executable robot actions.

\subsection{Real-World Experiments}

\begin{table}[!thb]
\centering
\caption{
Real-world results on 4 base tasks, with 30 trials per task.
}
\label{tab:real_world_main}
\scriptsize
\renewcommand{\arraystretch}{1.05}
\setlength{\tabcolsep}{2.2pt}
\begin{tabular}{@{}llcccccccc@{}}
\toprule
Group
& Method
& \makecell{Robot demos\\per task}
& \makecell{Human videos\\per task}
& \makecell{Basic\\Pick-Place}
& \makecell{Stack\\Paper Cups}
& \makecell{Insert\\Flowers}
& \makecell{Disassemble\\Blocks}
& Avg.
& \makecell{Inference\\latency (ms)} \\
\midrule
\multirow{2}{*}{\makecell[l]{Baselines}}
& $\pi_0$~\citep{black2026pi0visionlanguageactionflowmodel}
& 200 & 0 & 25/30 & 15/30 & 12/30 & 10/30 & 51.7 & 170 \\
& $\pi_{0.5}$~\citep{intelligence2025pi05visionlanguageactionmodelopenworld}
& 200 & 0 & 26/30 & 15/30 & 16/30 & 13/30 & 58.3 & 190 \\
\midrule
\multirow{3}{*}{\makecell[l]{Ours\\50 demos}}
& WALA
& 50 & 0 & 26/30 & 18/30 & 17/30 & 15/30 & 63.3 & \best{70} \\
& WALA
& 50 & 150 & 28/30 & 21/30 & 18/30 & 17/30 & 70.0 & \best{70} \\
& WALA
& 50 & 400 & \best{30/30} & 23/30
& 19/30 & 17/30
& 74.2 & \best{70} \\
\midrule
\multirow{2}{*}{\makecell[l]{Ours\\200 demos}}
& WALA
& 200 & 0 & 28/30 & 21/30 & 21/30 & 20/30 & 75.0 & \best{70} \\
& WALA
& 200 & 400
& \best{30/30}
& \best{25/30}
& \best{24/30}
& \best{21/30}
& \best{83.3}
& \best{70} \\
\bottomrule
\end{tabular}
\end{table}

\textbf{Experimental setup.}
We evaluate on an ARX robot. Policies for base-task comparisons are
jointly trained on only the 4 base tasks (Table~\ref{tab:real_world_main}),
whereas transfer policies are jointly trained using the available data from a 10-task set
(Table~\ref{tab:real_transfer}). Additional details on task co-training and evaluation are
provided in Appendix~\ref{sec:real_protocol}.

\textbf{Performance and inference latency.}
With 200 demonstrations and no human videos per base task, WALA reaches
75.0\% average base-task success, versus 51.7\% for
$\pi_0$ and 58.3\% for $\pi_{0.5}$
(Table~\ref{tab:real_world_main}). Even with 50 demonstrations, WALA
exceeds both baselines trained on 200. Pretrained interaction knowledge
and downstream future-prediction supervision help the policy anticipate object motion
and state changes, while depth supervision supports precise spatial
alignment. These complementary cues help explain the larger gains over
$\pi_{0.5}$ on stacking, insertion, and disassembly than on basic pick-place.
All inference latencies are measured per prediction on the same RTX 4090.
WALA takes 70\,ms using only the policy backbone and action head,
achieving roughly a $2.5\times$ inference speedup over
$\pi_0$ and $\pi_{0.5}$.

\begin{figure}[!t]
    \centering
    \includegraphics[width=\linewidth]{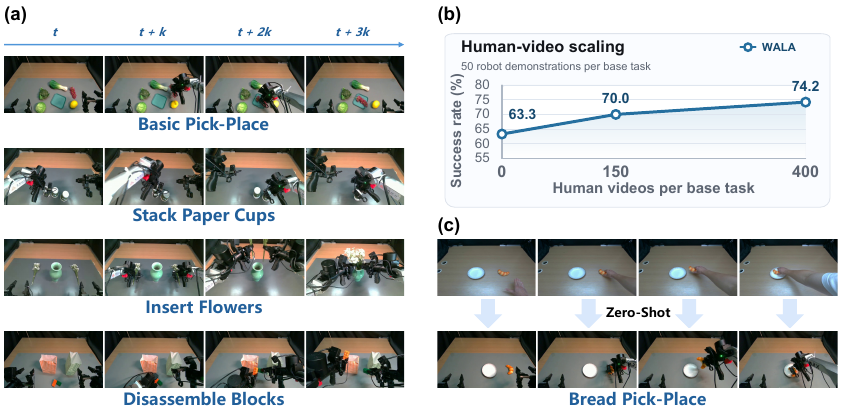}
    \caption{Real-world experiments. (a) Successful rollouts on the 4 base
    tasks. (b) Human-video scaling with
    50 robot demonstrations per base task. (c) Target-task zero-shot transfer
    to Bread Pick-Place using action-free egocentric human videos and no
    target-task robot demonstrations.}
    \label{fig:real_world_summary}
\end{figure}

\textbf{Human-video scaling.}
With 50 demonstrations per base task, adding human videos steadily raises
success from 63.3\% to 74.2\% (Fig.~\ref{fig:real_world_summary}(b)),
nearly matching 75.0\% with 4 times as many robot demonstrations per
base task. Human videos also improve performance with 200 demonstrations,
reaching 83.3\%. Additional human videos from similar scenes broaden
the coverage of object positions and manipulation patterns, helping
the policy handle more varied task configurations without requiring
additional robot demonstrations.

\begin{wraptable}{R}{.48\textwidth}
    \centering
    \caption{Transfer results as successful trials out of 30.
    Human-video counts are per target task.}
    \label{tab:real_transfer}
    \fontsize{8}{10}\selectfont
    \renewcommand{\arraystretch}{1.06}
    \setlength{\tabcolsep}{3pt}
    \begin{tabularx}{\linewidth}{@{}>{\raggedright\arraybackslash}Xcc@{}}
    \toprule
    \multirow{2}{*}{Task}
    & \multicolumn{2}{c}{Human videos} \\
    \cmidrule(lr){2-3}
    & 0 & 150 \\
    \midrule
    \multicolumn{3}{@{}l@{}}{\textit{Few-shot}\enspace (5 robot demos/task)} \\
    \addlinespace[2pt]
    Turn On Lamp
    & 13/30 & \best{22/30} \\
    Open Shelf-Unit Door
    & 8/30 & \best{19/30} \\
    Pen Pick-Place + Door Closing
    & 6/30 & \best{18/30} \\
    \midrule
    \multicolumn{3}{@{}l@{}}{\textit{Zero-shot}\enspace (0 robot demos/task)} \\
    \addlinespace[2pt]
    Bread Pick-Place
    & 9/30 & \best{24/30} \\
    Open Wooden Cabinet Door
    & 4/30 & \best{15/30} \\
    Place Two Apples on Plate
    & 0/30 & \best{8/30} \\
    \bottomrule
    \end{tabularx}
\end{wraptable}
\textbf{Target-task zero-shot transfer.}
With no target-task robot demonstrations, human videos raise
success across all 3 tasks (Table~\ref{tab:real_transfer}).
Transfer is strong for single-object pick-place
(Fig.~\ref{fig:real_world_summary}(c)) and cabinet opening,
while repeated 2-object placement remains more challenging.
Target-task zero-shot performance may also benefit from action
supervision on other co-trained robot tasks. Human videos help
the policy understand how new target objects are manipulated,
guiding the reuse of related action patterns in new interactions
rather than learning new action primitives from video alone.

\textbf{Few-shot transfer.}
Human videos also improve few-shot performance in switch operation,
door opening, and combined pick-place and door closing.
The gains extend beyond isolated grasping to object-state
changes and sequential manipulation. A few robot demonstrations provide task-specific action guidance,
while human videos supplement these limited examples with variations
in object placement and motion during task execution.

\WFclear
\section{Conclusion and Limitations}

WALA connects interaction learning from action-free videos with robot
control through a shared latent action representation. Semantic-geometric
delta modeling emphasizes interaction changes while retaining spatial
detail. Beyond matching pretrained targets, joint world and action
supervision refines this representation through downstream prediction
and grounds it in executable actions. Latent actions thus connect
interaction prediction with action generation. Co-training thereby
combines broader interaction experience from action-free videos with
the action grounding supplied by robot demonstrations. Experiments
show strong performance and data efficiency across simulation and
real robots, with task-relevant human videos further improving
manipulation and supporting target-task zero-shot and few-shot transfer.

Due to computational constraints, we have not systematically explored
scaling the LAM's pretraining data and model size. Large-scale robot-policy
pretraining on both action-labeled demonstrations and action-free
videos also remains unexplored. Investigating these directions could
further improve generalization across tasks, environments, and embodiments.

\clearpage
\subsection*{AI use statement}

We used generative AI tools for language polishing, literature
search, and drafting sections of the manuscript. The authors
reviewed all AI-assisted content and take responsibility for
the final paper.

\subsection*{Ethics statement}

This work studies robot manipulation using publicly available
datasets, robot demonstrations, and egocentric human videos.
The public datasets were used in accordance with their
applicable licenses and terms of use. The human videos
collected for this study depict task-relevant object
manipulation and were recorded with the participants'
informed consent. Real-robot experiments were conducted
under human supervision.

\subsection*{Reproducibility statement}

The model and training objectives are described in Section 3,
and the experimental settings and evaluation protocols in Section 4.
Appendix Table~\ref{tab:pretraining_data} summarizes the pretraining
data composition, and Appendix~\ref{sec:model_training_configuration}
provides the model and training configurations.
Per-task simulation results are reported in Appendix
Tables~\ref{tab:appendix_robotwin_tasks}
and~\ref{tab:appendix_robocasa_tasks}.
The project website, code, and model weights are available at
\url{https://liujiahao2077.github.io/WALA.github.io}.

%
%

\setlength{\bibsep}{1pt plus 0.5pt}
\bibliography{iclr2027_conference}
\bibliographystyle{iclr2027_conference}

\clearpage
\appendix
\section{Additional Experimental Details}

\subsection{Latent Action Pretraining Data}
\label{sec:pretraining_data}

The LAM is pretrained for 400,000 optimization steps using a
batch size of 1,024. Table~\ref{tab:pretraining_data} summarizes the video data
used for pretraining. Each video sample contains one current frame and four
evenly sampled future frames. For normal-speed videos, the sampled clip
spans approximately 1.6 seconds. For some human videos with fast hand
motions, we use shorter windows to avoid covering excessive manipulation
progress within a single clip. This temporal adaptation applies to the LAM
pretraining corpus; sampling during robot policy training is described in
Appendix~\ref{sec:video_policy_inputs}.

The RoboCasa and RoboTwin videos used for LAM pretraining come exclusively
from their training sets. No evaluation rollouts or test-set videos are
included, and pretraining does not use action annotations. In the 10\%
robot-demonstration setting, the LAM has already seen videos from the full
training set. Thus, 10\% denotes the action-supervision budget during policy
training, not the amount of video available to LAM pretraining.

\begin{table}[ht]
\centering
\caption{Composition of the video corpus used to pretrain the
semantic-geometric latent action model.}
\label{tab:pretraining_data}
\small
\renewcommand{\arraystretch}{1.08}
\setlength{\tabcolsep}{9.8pt}
\begin{tabular}{llr}
\toprule
Data category & Dataset & Hours \\
\midrule
\multirow{8}{*}{Human interaction videos}
& SSv2~\citep{goyal2017something} & 240 \\
& Ego-Exo4D~\citep{grauman2024egoexo4d} & 1,286 \\
& Xperience-10M~\citep{ropedia2026xperience10m} & 921 \\
& Egocentric-10K~\citep{buildai2025egocentric10k} & 10,000 \\
& Ego4D~\citep{grauman2022ego4d} & 3,670 \\
& EgoVerse~\citep{punamiya2026egoverse} & 1,362 \\
& EgoDex~\citep{hoque2025egodexlearningdexterousmanipulation} & 830 \\
\cmidrule(lr){2-3}
& \textit{Subtotal} & \textit{18,309} \\
\midrule
\multirow{4}{*}{Real-robot data}
& RoboCOIN~\citep{wu2025robocoinopensourcedbimanualrobotic} & 704 \\
& AgiBot World 2026~\citep{agibotworld2026dataset} & 433 \\
& AgiBot World Beta~\citep{agibotworld2025colosseo} & 2,007 \\
\cmidrule(lr){2-3}
& \textit{Subtotal} & \textit{3,144} \\
\midrule
\multirow{4}{*}{Simulated robot data}
& InternData-A1~\citep{cai2026internvlaa1unifyingunderstandinggeneration} & 3,656 \\
& RoboCasa~\citep{nasiriany2024robocasalargescalesimulationeveryday} & 81 \\
& RoboTwin~\citep{chen2025robotwin20scalabledata} & 34 \\
\cmidrule(lr){2-3}
& \textit{Subtotal} & \textit{3,771} \\
\midrule
\multicolumn{2}{l}{\textbf{Total}} & \textbf{25,224} \\
\bottomrule
\end{tabular}
\end{table}

\subsection{Policy Inputs and Video Sampling}
\label{sec:video_policy_inputs}

\paragraph{Observations and instructions.}
The policy normally receives a primary view and two wrist views, whereas
the LAM uses only the primary view. For egocentric human videos, the two
missing wrist views are represented by learnable embeddings, with text
indicating which views are missing.
Robot state is omitted when unavailable. For each task scene in our
self-collected human videos, a vision-language model generates 20
semantically equivalent instructions. Human videos and the corresponding
robot demonstrations share this instruction set, from which an instruction
is sampled randomly during training. These videos have no action
annotations; language instructions provide task context, not motion targets.

\paragraph{Temporal sampling.}
Real-robot policy training uses our self-collected human videos. We match the
manipulation pace to robot demonstrations during recording, or adjust the
video speed afterward when human videos are collected at a faster pace.
Human and robot clips are then sampled over the same temporal duration,
corresponding to one robot action chunk. Each clip contains one current
frame and four future frames at evenly spaced offsets. This duration
matching does not require frame-aligned human and robot demonstrations.

\paragraph{Mixed policy training.}
When co-training with human videos and robot demonstrations, the two data
sources are sampled with weights of 0.5 and 0.5, respectively. These are
sampling weights rather than a fixed equal count in every minibatch.
Encoder distillation and decoder prediction apply to both sources, while
the action loss applies only to samples with robot action annotations.
For RoboCasa video scaling, additional robot videos are used without
action supervision and receive both LAM losses.
The loss reductions are specified in Appendix~\ref{sec:loss_implementation}.

\subsection{Model and Training Configuration}
\label{sec:model_training_configuration}

Tables~\ref{tab:model_configuration} and~\ref{tab:training_configuration}
summarize the architecture and optimization settings. WALA and WALA-Wan share
the semantic-geometric LAM and its training losses; their policy-backbone
interfaces are compared in Fig.~\ref{fig:wala_backbone_variants}.

\paragraph{Latent action adapters.}
The encoder and decoder adapters have the same architecture. Each first projects the $H$ policy tokens to dimension 1,024
and adds learned temporal embeddings. A one-layer, eight-head Transformer
decoder uses 32 learned queries to attend to these tokens and returns
32 tokens of dimension 1,024. Thus, each adapter maps
$\mathbb{R}^{H\times d_{\mathrm{policy}}}$ to
$\mathbb{R}^{32\times1024}$, where $d_{\mathrm{policy}}$ is the policy token
dimension.

\begin{table}[ht]
\centering
\caption{Core model settings for WALA and WALA-Wan on RoboTwin.}
\label{tab:model_configuration}
\small
\renewcommand{\arraystretch}{1.08}
\setlength{\tabcolsep}{5pt}
\begin{minipage}{0.92\textwidth}
\raggedright
\textit{Policy-backbone settings}\par\vspace{2pt}

\begin{tabular}{@{}
    >{\raggedright\arraybackslash}p{0.27\linewidth}
    >{\raggedright\arraybackslash}p{0.31\linewidth}
    >{\raggedright\arraybackslash}p{\dimexpr0.42\linewidth-4\tabcolsep\relax}
    @{}}
\toprule
Component & Standard WALA & WALA-Wan \\
\midrule
Policy backbone & Qwen3-VL-4B-Instruct & Wan2.2-TI2V-5B \\
Policy input resolution & $256\times256$ & $384\times320$ \\
Action head & MLP (hidden size 2,560) & MLP (hidden size 3,072) \\
Latent action generation & Learned action queries in the policy backbone & 2-layer, 24-head action-query resampler over backbone hidden tokens \\
\bottomrule
\end{tabular}

\vspace{7pt}
\textit{Shared semantic-geometric latent action model}\par\vspace{2pt}

\begin{tabular}{@{}
    >{\raggedright\arraybackslash}p{0.41\linewidth}
    >{\raggedright\arraybackslash}p{\dimexpr0.59\linewidth-2\tabcolsep\relax}
    @{}}
\toprule
Component & Configuration \\
\midrule
Semantic feature encoder & DINOv3 ViT-B/16; feature dimension 768 \\
Depth estimator & Depth Anything V2-L; input resolution $224\times224$ \\
Latent action representation & 32 latent action tokens; hidden dimension 1,024 \\
LAM encoder and decoder & 8 encoder layers; 4 decoder layers; 8 attention heads \\
\bottomrule
\end{tabular}
\end{minipage}
\end{table}

The essential optimization settings are summarized in
Table~\ref{tab:training_configuration}. Batch size denotes the effective total
batch size across devices. All stages use AdamW with
$(\beta_1,\beta_2)=(0.9,0.95)$ and $\epsilon=10^{-8}$, cosine learning-rate
decay with a minimum learning rate of $5\times10^{-7}$, gradient clipping at
1.0, and random seed 42 unless otherwise specified.

\begin{table}[ht]
\centering
\caption{Essential training settings for LAM pretraining and policy training.}
\label{tab:training_configuration}
\small
\renewcommand{\arraystretch}{1.08}
\setlength{\tabcolsep}{4pt}
\begin{tabular*}{\textwidth}{@{\extracolsep{\fill}}lcccc@{}}
\toprule
Setting & \makecell{LAM\\pretraining} & RoboTwin & RoboCasa & \makecell{Real\\robot} \\
\midrule
Training steps & 400,000 & 200,000 & 100,000 & 50,000 \\
Effective batch size & 1,024 & 192 & 256 & 256 \\
Action dimension & -- & 14 & 47 & 14 \\
Action horizon $H$ & -- & 50 & 32 & 50 \\
Default learning rate & $1\times10^{-4}$ & $1\times10^{-5}$ & $3\times10^{-5}$ & $1\times10^{-5}$ \\
Policy-backbone learning rate & -- & $1\times10^{-5}$ & $1\times10^{-5}$ & $1\times10^{-5}$ \\
Decoder learning rate & $1\times10^{-4}$ & $1\times10^{-5}$ & $1\times10^{-5}$ & $1\times10^{-5}$ \\
Action head / adapters learning rate & -- & $1\times10^{-4}$ & $1\times10^{-4}$ & $1\times10^{-4}$ \\
Robot-state dropout & -- & 0 & 0 & 0.1 \\
\bottomrule
\end{tabular*}
\end{table}

For WALA-Wan on RoboTwin, we use the same effective batch size of 192 and
200,000 training steps as standard WALA. The Wan backbone uses a learning rate
of $1\times10^{-5}$, while the action-query resampler uses
$1\times10^{-4}$. The Wan VAE and text encoder remain frozen during training.

During Stage I, the trainable LAM modules are initialized from scratch, while
DINOv3 and Depth Anything V2 remain frozen. Standard policy training on RoboTwin and RoboCasa-GR1-Tabletop
uses the pretrained LAM. Real-robot policy training is initialized from a complete
RoboTwin-trained WALA checkpoint to accelerate convergence; this initialization
is not large-scale robot-policy pretraining.

\paragraph{Ablation configurations.}
All variants in Table~\ref{tab:ablation} retain robot action supervision.
Removing distillation or prediction supervision leaves the remaining
loss weights unchanged. The semantic-only variant uses a LAM whose encoder
does not encode depth and whose decoder does not predict depth. The
semantic-only and semantic-geometric LAMs have different architectures
and are pretrained in separate runs. The variant without LAM pretraining
initializes its decoder randomly, while retaining the pretrained policy
backbone and frozen visual feature extractors.

Training and policy inference for RoboTwin and RoboCasa-GR1-Tabletop are
performed on Alibaba Cloud's PPU platform. NVIDIA H20 GPUs are used for
RoboTwin environment rendering.

\paragraph{LAM pretraining resources and reuse.}
The semantic-geometric LAM contains approximately 400 million parameters.
Pretraining on the 25,224-hour video corpus for 400,000 optimization steps
with a global batch size of 1,024 takes approximately 200 hours on
128 Alibaba Cloud PPU accelerators, corresponding to approximately
25,600 PPU-hours for this pretraining run. The pretrained LAM is reused
across downstream tasks without repeating large-scale video pretraining.
During robot policy training, task-relevant egocentric human videos
provide additional supervision without action annotations, improving
manipulation performance with limited robot demonstrations.

\paragraph{LAM-based baseline reproductions on RoboCasa.}
We adapt the released Moto~\citep{chen2025motolatentmotiontoken} and
UniVLA~\citep{bu2025univlalearningacttaskcentric} implementations to
RoboCasa-GR1-Tabletop. For a fair comparison on the target benchmarks,
the Moto and UniVLA results in Table~\ref{tab:robocasa_main} use LAM
checkpoints fine-tuned on RoboTwin and RoboCasa. These checkpoints are
also used for the action probes and encoder-attention visualizations.
Both use the same training set as WALA, with 1,000
demonstrations for each of the 24 tasks, a global batch size of 256,
a training budget of 100,000 optimizer updates, and 32-step action chunks.
Each baseline is trained on 16 Alibaba Cloud PPU accelerators, using the
same accelerator platform as WALA. Evaluation follows the same benchmark
protocol as WALA, with 50 episodes per task, 32 actions executed per policy
prediction, a limit of 720 simulator steps per episode, and the same
simulator-defined success criterion. Moto and UniVLA achieve 547/1200
(45.58\%) and 614/1200 (51.17\%), respectively. Results are rounded to one
decimal place in Table~\ref{tab:robocasa_main}.

\subsection{Loss Implementation}
\label{sec:loss_implementation}

The regression notation in the Method distinguishes the
semantic and geometric terms without prescribing the same penalty for both.
Semantic-feature regression uses L1, whereas depth-map regression uses
Smooth L1,
\begin{equation}
\begin{aligned}
\mathcal{L}_{\mathrm{reg}}^{\mathrm{sem}}
    &= \operatorname{mean}\!\left|\widehat{\Delta X}-\Delta X\right|,\\
\mathcal{L}_{\mathrm{reg}}^{\mathrm{dep}}
    &= \operatorname{SmoothL1}(\widehat{\Delta D},\Delta D).
\end{aligned}
\end{equation}
The semantic cosine term $\mathcal{L}_{\mathrm{cos}}$ is one minus the
cosine similarity between predicted and target feature-delta vectors,
averaged over samples, future steps, and spatial locations. It complements
element-wise regression by matching the direction of feature change.

The depth-gradient term compares horizontal and vertical finite differences,
\begin{equation}
\mathcal{L}_{\mathrm{grad}}
=\sum_{d\in\{x,y\}}
\operatorname{mean}\!\left|
\nabla_d\widehat{\Delta D}-\nabla_d\Delta D\right|.
\end{equation}
This term encourages more accurate recovery of local depth variations.
All four prediction terms are evaluated in delta space. The depth maps
used by the LAM are estimated at $224\times224$ resolution, and the decoder
predicts depth deltas at this resolution. Before depth deltas
are formed, the current and future maps share a clip-wise normalization:
log-depth values are centered by their median, scaled by their median
absolute deviation, and clipped to $[-3,3]$. The depth-map notation in the
Method refers to these normalized estimates. Sharing the normalization
statistics puts all frames on a common scale within the clip, but does not
remove errors or temporal inconsistencies in monocular depth estimation.

For latent-action distillation in Eq.~(\ref{eq:distillation}), let
$U=(u_1,\ldots,u_{N_z})$ and $V=(v_1,\ldots,v_{N_z})$ denote the adapted
policy tokens and encoder targets, respectively, with $N_z=32$. Then
\begin{equation}
\ell_{\mathrm{cos}}(U,V)
=\frac{1}{N_z}\sum_{j=1}^{N_z}
\left(1-\frac{\langle u_j,v_j\rangle}
{\|u_j\|_2\|v_j\|_2}\right).
\end{equation}
Layer normalization is applied independently over each token's feature
dimension without learned affine parameters. Smooth L1 is averaged over
tokens and feature dimensions, with transition threshold 1. Encoder targets
receive no gradient. We set
$\lambda_{\mathrm{cos}}=0.1$, $\lambda_{\mathrm{grad}}=0.2$, and
$\lambda_{\mathrm{dep}}=1.0$ for LAM pretraining. The policy's world-model
loss uses the same prediction terms and weights, with
$\lambda_{\mathrm{distill}}=0.005$ and $\lambda_{\mathrm{wm}}=0.5$ in the
overall policy objective.

\paragraph{Loss-weight selection.}
We choose the policy-loss weights empirically based on the magnitudes of
the separately averaged loss terms during training. The action-loss
coefficient is fixed at 1. The auxiliary weights are selected to balance
distillation and world-model supervision without overwhelming the
primary action-prediction objective, and remain fixed during training.
These auxiliary terms also provide representation-level regularization
by constraining policy-generated latent actions through distillation
and future prediction.
Decoder prediction compares the policy tokens' predicted outcomes with
the semantic-feature and depth-map deltas in downstream videos, rather
than requiring them only to reproduce the frozen encoder's targets.
The action loss jointly links the policy tokens to recorded robot actions.

\paragraph{Minibatch reduction.}
Let $\mathcal{B}$ denote a policy-training minibatch and
$\mathcal{B}_{\mathrm{lab}}$ its samples with robot action annotations.
For a labeled sample $i$, the action loss is
\begin{equation}
\mathcal{L}_{\mathrm{act},i}
=\frac{1}{H d_a}\sum_{h=0}^{H-1}
\left\|\widehat a_{i,t+h}-a_{i,t+h}\right\|_1,
\end{equation}
where $d_a$ is the action dimension and $H$ is the chunk length.
This loss is averaged over $\mathcal{B}_{\mathrm{lab}}$; distillation and
world-model losses are averaged over all of $\mathcal{B}$,
\begin{equation}
\begin{aligned}
\mathcal{L}_{\mathrm{policy}}^{\mathcal{B}}
&= \frac{1}{\max(1,|\mathcal{B}_{\mathrm{lab}}|)}
   \sum_{i\in\mathcal{B}_{\mathrm{lab}}}\mathcal{L}_{\mathrm{act},i}\\
&\quad + \frac{1}{|\mathcal{B}|}\sum_{i\in\mathcal{B}}
\left(\lambda_{\mathrm{distill}}\mathcal{L}_{\mathrm{distill},i}
+\lambda_{\mathrm{wm}}\mathcal{L}_{\mathrm{wm},i}\right).
\end{aligned}
\end{equation}
The action term is zero for an unlabeled minibatch. This separately normalized
reduction implements the masked objective in Sec.~\ref{sec:policy_training}
without including unlabeled samples in the action-loss denominator.

\begin{figure}[!htbp]
    \centering
    \includegraphics[width=\linewidth]{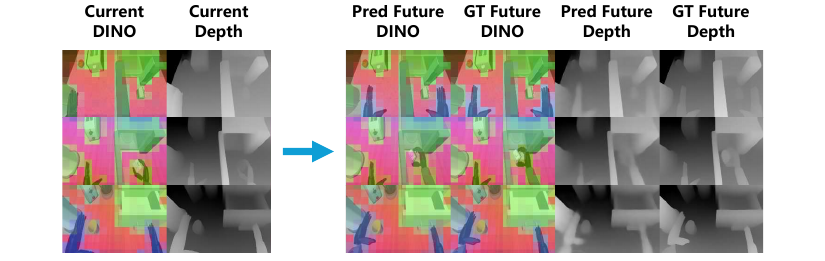}
    \caption{
    Future prediction from policy-generated latent action tokens. Through the
    decoder adapter, these tokens condition the LAM decoder to predict
    semantic-feature and depth-map deltas. Adding the deltas to the current
    representations yields the future predictions shown alongside their targets.
    DINOv3 features are PCA-projected to RGB and overlaid on the corresponding
    RGB frames; depth maps are shown directly.
    }
    \label{fig:pred_vis}
\end{figure}

\paragraph{Visualization.}
Figure~\ref{fig:wala_overview} displays the decoder's predicted signed
deltas. Neutral gray indicates little change; DINOv3 delta colors visualize
feature directions through PCA rather than fixed semantic categories.
Depth-delta brightness indicates signed changes in normalized depth estimates,
not metric distances. In Fig.~\ref{fig:pred_vis}, policy-generated latent action
tokens condition the LAM decoder through the decoder adapter. Adding its
predicted deltas to the current DINOv3 features and depth maps yields the
future predictions shown alongside their targets.
The DINOv3 feature maps are PCA-projected and overlaid on RGB frames, while
depth maps are shown directly.

\FloatBarrier
\subsection{Policy Backbone Variants}
\label{sec:policy_backbones}

\begin{figure*}[!htbp]
    \centering
    \includegraphics[width=0.96\textwidth]{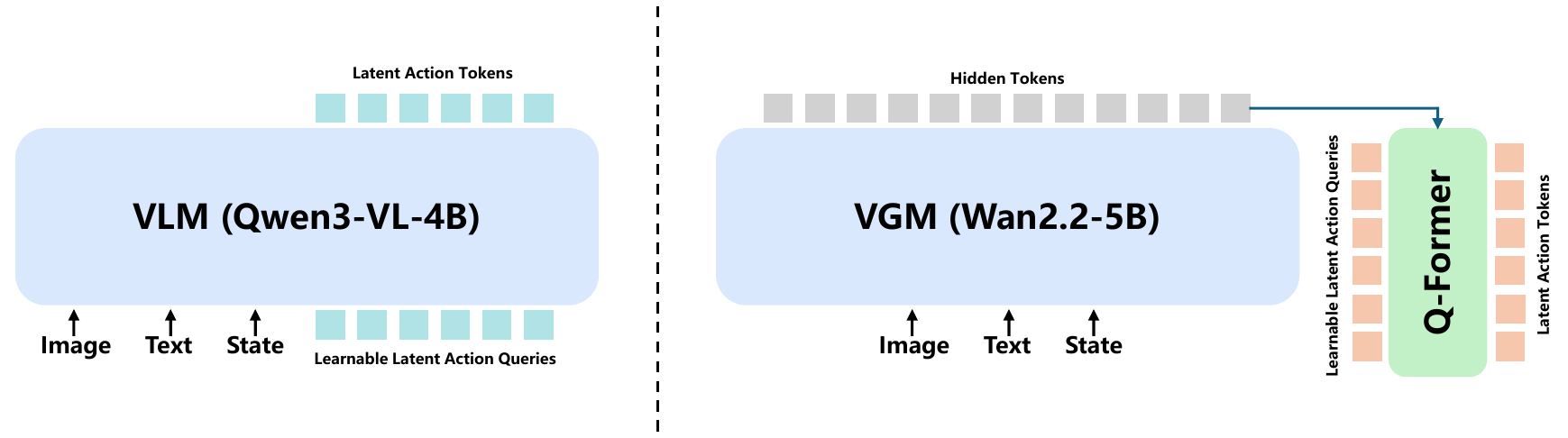}
    \caption{
    Comparison of the two WALA policy backbones. Standard WALA reads latent action
    tokens from learned action queries in Qwen3-VL-4B-Instruct, while WALA-Wan uses an
    action-query resampler over Wan2.2-TI2V-5B hidden tokens. Both variants
    use the same latent-action distillation, semantic-geometric world-model
    supervision, and robot-action prediction objectives.
    }
    \label{fig:wala_backbone_variants}
\end{figure*}

Figure~\ref{fig:wala_backbone_variants} compares the action-query interfaces
of the two backbones specified in Sec.~\ref{sec:experimental_setup}.
Both use the training objective in Sec.~\ref{sec:policy_training}.

\subsection{Real-Robot Experimental Protocol}
\label{sec:real_protocol}

For base-task comparisons, WALA, $\pi_0$, and $\pi_{0.5}$ each jointly
train a single policy on only the same 4 base tasks. The baselines use
the same 200 robot demonstrations per base task and evaluation protocol
as WALA's 200-demonstration setting.

For transfer experiments, each WALA data setting trains a single policy
in a joint 10-task setting comprising the 4 base tasks and 6 transfer tasks,
using the available robot demonstrations and human videos.
In both transfer conditions, each base task uses 200 robot demonstrations
and 400 human videos; only the target-task human-video budget varies.
In the condition without target-task human videos, the 3 zero-shot tasks have
neither robot demonstrations nor human videos and therefore contribute
no training samples.
All tasks are evaluated for 30 trials per task
(Tables~\ref{tab:real_world_main} and~\ref{tab:real_transfer}).
Average base-task success is computed over 120 trials.

The 3 zero-shot tasks use no target-task robot demonstrations, whereas
the 3 few-shot tasks use 5 per task; both compare zero versus 150
human videos per target task. Place Two Apples on Plate requires picking
and placing the apples sequentially. The 2 door-opening tasks use
different cabinet instances and scene layouts. Human recordings contain
only single-camera egocentric video, without measured or post-processed
end-effector (EEF) trajectories.

\begin{figure}[!htbp]
    \centering
    \includegraphics[width=\linewidth]{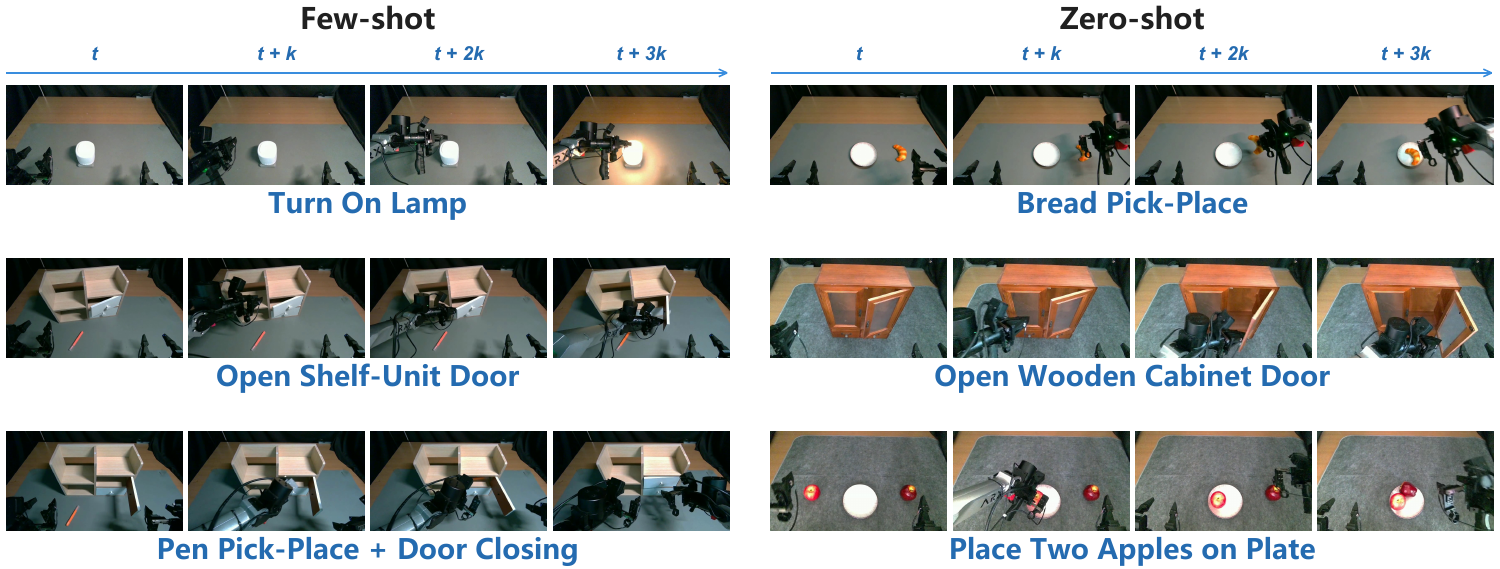}
    \caption{Real-robot transfer rollouts on the 3 few-shot tasks (left)
    and 3 zero-shot tasks (right). Each row shows 4 equally spaced frames
    spanning the start and completion of an operation.}
    \label{fig:real_transfer_sequences}
\end{figure}

Figure~\ref{fig:real_transfer_sequences} visualizes representative rollouts
for all 6 transfer tasks.

\subsection{Per-Task Benchmark Results}

Tables~\ref{tab:appendix_robotwin_tasks} and~\ref{tab:appendix_robocasa_tasks}
report individual task results. RoboTwin uses 100 evaluation episodes per
task in each setting, and RoboCasa-GR1-Tabletop uses 50 per task. Within each
benchmark, equal trial counts make the aggregate success rate equivalent to
the unweighted mean of per-task success rates.

\begingroup
\scriptsize
\renewcommand{\tablename}{\normalsize Table}
\renewcommand{\arraystretch}{1.02}
\setlength{\tabcolsep}{4pt}
\setlength{\LTcapwidth}{\textwidth}
\begin{longtable}{@{}
    >{\raggedright\arraybackslash}p{0.28\textwidth}
    *{4}{>{\centering\arraybackslash}p{0.13\textwidth}}
    @{}}
\caption{\normalsize
Per-task RoboTwin results for WALA and WALA-Wan, reported as
successful trials out of 100 evaluation episodes.
}
\label{tab:appendix_robotwin_tasks} \\
\toprule
\multirow{2}{*}{Task}
& \multicolumn{2}{c}{WALA}
& \multicolumn{2}{c}{WALA-Wan} \\
\cmidrule(lr){2-3}\cmidrule(lr){4-5}
& Clean & Random & Clean & Random \\
\midrule
\endfirsthead
\caption[]{\normalsize Per-task RoboTwin results (continued).} \\
\toprule
\multirow{2}{*}{Task}
& \multicolumn{2}{c}{WALA}
& \multicolumn{2}{c}{WALA-Wan} \\
\cmidrule(lr){2-3}\cmidrule(lr){4-5}
& Clean & Random & Clean & Random \\
\midrule
\endhead
\midrule
\multicolumn{5}{r}{\textit{Continued on the next page.}} \\
\endfoot
\bottomrule
\endlastfoot
adjust\_bottle             & 100/100 & 100/100 & 100/100 & 100/100 \\
beat\_block\_hammer        & 98/100  & 96/100  & 98/100  & 96/100 \\
blocks\_ranking\_rgb       & 99/100  & 98/100  & 100/100 & 98/100 \\
blocks\_ranking\_size      & 77/100  & 89/100  & 89/100  & 89/100 \\
click\_alarmclock          & 100/100 & 100/100 & 97/100  & 95/100 \\
click\_bell                & 100/100 & 99/100  & 98/100  & 96/100 \\
dump\_bin\_bigbin          & 95/100  & 99/100  & 99/100  & 97/100 \\
grab\_roller               & 100/100 & 100/100 & 100/100 & 100/100 \\
handover\_block            & 86/100  & 88/100  & 92/100  & 79/100 \\
handover\_mic              & 100/100 & 100/100 & 98/100  & 100/100 \\
hanging\_mug               & 59/100  & 53/100  & 54/100  & 49/100 \\
lift\_pot                  & 100/100 & 100/100 & 100/100 & 100/100 \\
move\_can\_pot             & 84/100  & 87/100  & 87/100  & 90/100 \\
move\_pillbottle\_pad      & 100/100 & 100/100 & 97/100  & 99/100 \\
move\_playingcard\_away    & 100/100 & 100/100 & 99/100  & 98/100 \\
move\_stapler\_pad         & 59/100  & 63/100  & 84/100  & 77/100 \\
open\_laptop               & 100/100 & 100/100 & 98/100  & 100/100 \\
open\_microwave            & 89/100  & 99/100  & 88/100  & 94/100 \\
pick\_diverse\_bottles     & 67/100  & 80/100  & 76/100  & 72/100 \\
pick\_dual\_bottles        & 91/100  & 93/100  & 90/100  & 88/100 \\
place\_a2b\_left           & 91/100  & 99/100  & 93/100  & 94/100 \\
place\_a2b\_right          & 94/100  & 95/100  & 96/100  & 94/100 \\
place\_bread\_basket       & 96/100  & 93/100  & 95/100  & 92/100 \\
place\_bread\_skillet      & 97/100  & 89/100  & 88/100  & 93/100 \\
place\_burger\_fries       & 98/100  & 100/100 & 99/100  & 99/100 \\
place\_can\_basket          & 84/100  & 79/100  & 80/100  & 83/100 \\
place\_cans\_plasticbox     & 98/100  & 100/100 & 98/100  & 99/100 \\
place\_container\_plate     & 100/100 & 100/100 & 99/100  & 97/100 \\
place\_dual\_shoes          & 81/100  & 82/100  & 83/100  & 96/100 \\
place\_empty\_cup           & 99/100  & 100/100 & 99/100  & 100/100 \\
place\_fan                  & 91/100  & 96/100  & 98/100  & 94/100 \\
place\_mouse\_pad           & 74/100  & 85/100  & 93/100  & 88/100 \\
place\_object\_basket       & 90/100  & 87/100  & 90/100  & 87/100 \\
place\_object\_scale        & 97/100  & 100/100 & 96/100  & 99/100 \\
place\_object\_stand        & 98/100  & 98/100  & 91/100  & 94/100 \\
place\_phone\_stand         & 96/100  & 99/100  & 98/100  & 95/100 \\
place\_shoe                 & 99/100  & 100/100 & 98/100  & 99/100 \\
press\_stapler              & 99/100  & 100/100 & 97/100  & 99/100 \\
put\_bottles\_dustbin       & 73/100  & 82/100  & 87/100  & 85/100 \\
put\_object\_cabinet        & 89/100  & 90/100  & 89/100  & 100/100 \\
rotate\_qrcode              & 84/100  & 82/100  & 83/100  & 87/100 \\
scan\_object                & 88/100  & 93/100  & 92/100  & 89/100 \\
shake\_bottle               & 100/100 & 100/100 & 100/100 & 99/100 \\
shake\_bottle\_horizontally & 100/100 & 100/100 & 100/100 & 99/100 \\
stack\_blocks\_three        & 94/100  & 98/100  & 98/100  & 99/100 \\
stack\_blocks\_two          & 100/100 & 100/100 & 100/100 & 99/100 \\
stack\_bowls\_three         & 84/100  & 85/100  & 84/100  & 85/100 \\
stack\_bowls\_two           & 96/100  & 100/100 & 98/100  & 91/100 \\
stamp\_seal                 & 88/100  & 88/100  & 82/100  & 92/100 \\
turn\_switch                & 49/100  & 76/100  & 70/100  & 71/100 \\
\midrule
\multicolumn{5}{@{}p{\dimexpr0.80\textwidth+8\tabcolsep\relax}@{}}{\textit{WALA overall:} 4531/5000 (90.62\%) on Clean and 4640/5000 (92.80\%) on Random; 9171/10000 (91.71\%) combined.} \\
\multicolumn{5}{@{}p{\dimexpr0.80\textwidth+8\tabcolsep\relax}@{}}{\textit{WALA-Wan overall:} 4618/5000 (92.36\%) on Clean and 4615/5000 (92.30\%) on Random; 9233/10000 (92.33\%) combined.} \\
\end{longtable}
\endgroup

\begin{table*}[!htbp]
\centering
\caption{
Per-task WALA results on RoboCasa-GR1-Tabletop. Results show successful
trials out of 50 evaluation episodes and the corresponding success rates.
}
\label{tab:appendix_robocasa_tasks}
\scriptsize
\renewcommand{\arraystretch}{1.06}
\setlength{\tabcolsep}{3.1pt}
\begin{tabular}{@{}r l c c@{\hspace{0.35cm}}r l c c@{}}
\toprule
\# & Task & Success & Rate & \# & Task & Success & Rate \\
\midrule
1  & PnPBottleToCabinetClose      & 41/50 & 82.0\% & 13 & PlacematToBowl          & 37/50 & 74.0\% \\
2  & PnPCanToDrawerClose          & 48/50 & 96.0\% & 14 & PlacematToPlate         & 33/50 & 66.0\% \\
3  & PnPCupToDrawerClose          & 43/50 & 86.0\% & 15 & PlacematToTieredshelf   & 21/50 & 42.0\% \\
4  & PnPMilkToMicrowaveClose      & 39/50 & 78.0\% & 16 & PlateToBowl             & 36/50 & 72.0\% \\
5  & PnPPotatoToMicrowaveClose    & 39/50 & 78.0\% & 17 & PlateToCardboardbox     & 25/50 & 50.0\% \\
6  & PnPWineToCabinetClose        & 31/50 & 62.0\% & 18 & PlateToPan              & 26/50 & 52.0\% \\
7  & CuttingboardToBasket         & 43/50 & 86.0\% & 19 & PlateToPlate            & 46/50 & 92.0\% \\
8  & CuttingboardToCardboardbox   & 33/50 & 66.0\% & 20 & TrayToCardboardbox      & 43/50 & 86.0\% \\
9  & CuttingboardToPan            & 47/50 & 94.0\% & 21 & TrayToPlate             & 49/50 & 98.0\% \\
10 & CuttingboardToPot            & 40/50 & 80.0\% & 22 & TrayToPot               & 42/50 & 84.0\% \\
11 & CuttingboardToTieredbasket   & 25/50 & 50.0\% & 23 & TrayToTieredbasket      & 40/50 & 80.0\% \\
12 & PlacematToBasket             & 48/50 & 96.0\% & 24 & TrayToTieredshelf       & 27/50 & 54.0\% \\
\midrule
\multicolumn{8}{@{}l@{}}{\textit{Overall:} 902/1200 (75.17\%).} \\
\bottomrule
\end{tabular}
\end{table*}

\FloatBarrier
\section{Additional Analyses of Latent Actions}
\label{sec:latent_action_analyses}

\subsection{Latent Action Retrieval}
\label{sec:latent_retrieval}

\begin{figure}[!htbp]
    \centering
    \includegraphics[width=\linewidth]{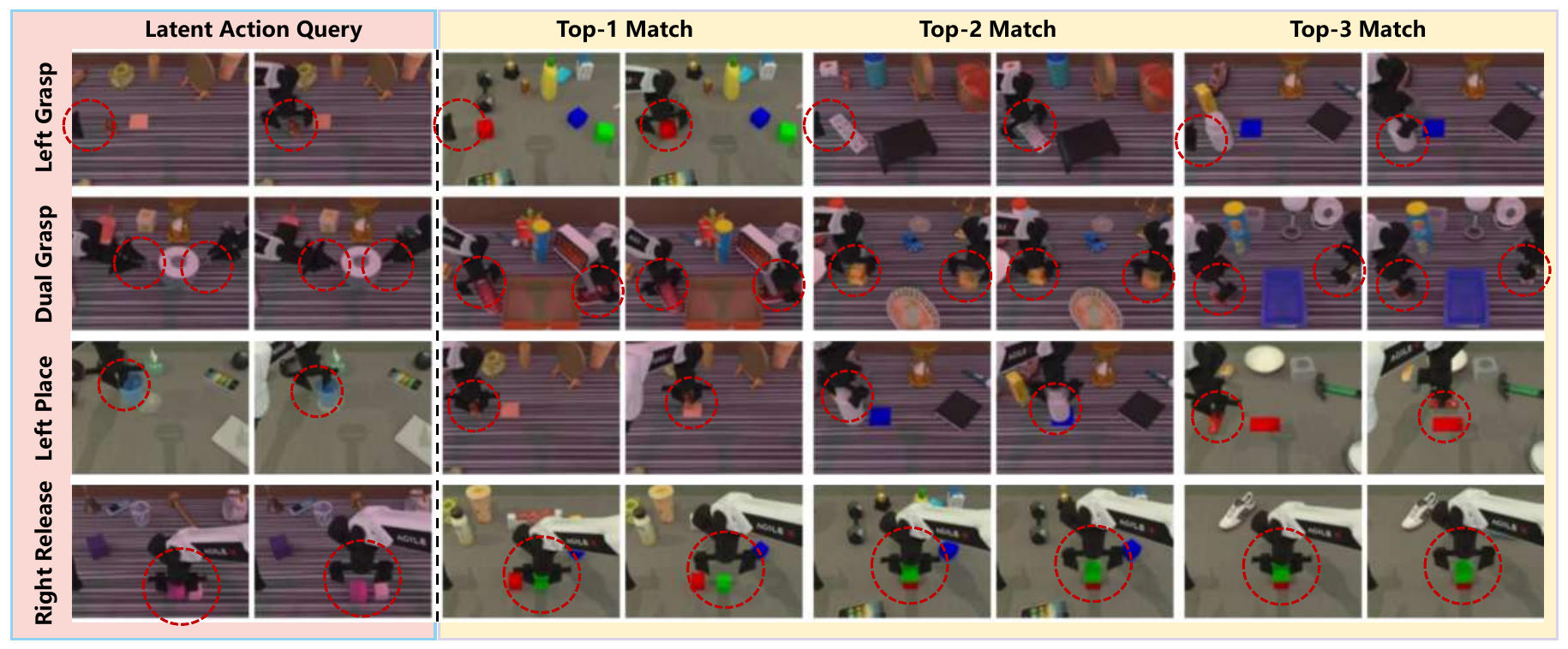}
    \caption{
    Retrieval using LAM-encoded latent actions. Example queries retrieve similar manipulation
    primitives across different objects, layouts, and backgrounds.
    }
    \label{fig:latent_retrieval}
\end{figure}

Figure~\ref{fig:latent_retrieval} illustrates retrieval using latent actions
encoded by the pretrained LAM, recovering similar manipulation primitives
across different objects, layouts, and backgrounds.

\subsection{Decoder Dependence on Policy-Generated Tokens}
\label{sec:token_dependency}

\paragraph{Protocol.}
We examine whether the semantic-geometric decoder uses the latent action
tokens produced by the RoboCasa policy. We freeze the jointly trained
policy and decoder for this diagnostic. We select 240 clips, one
from each of ten demonstration episodes for each of the 24 tasks, and
predict semantic-feature and depth-map deltas at future frame offsets
8, 16, 24, and 32. This is an offline diagnostic on the benchmark's
training-distribution demonstrations.
For each clip, the current visual features and normalized depth are held
fixed. We keep the original tokens, replace them with tokens from a
different episode of the same task or from another task, or set them to
zero. Replacements are made after the decoder adapter and preserve the
token sequence shape. Each swap condition averages three independently
sampled donors per recipient; these are repeated interventions on one
trained model, not independently trained policies.

\begin{table}[htbp]
\centering
\caption{Decoder token replacements on RoboCasa-GR1-Tabletop.
Errors are averaged over 240 clips; lower is better.
The last column reports the relative increase in the world-model loss
compared with original tokens.}
\label{tab:token_dependency}
\small
\renewcommand{\arraystretch}{1.12}
\setlength{\tabcolsep}{7pt}
\begin{tabular}{lcccc}
\toprule
Decoder tokens & \makecell{Semantic delta\\MSE $\downarrow$}
& \makecell{Depth delta\\MSE $\downarrow$}
& $\mathcal{L}_{\mathrm{wm}}\!\downarrow$
& \makecell{Loss\\increase} \\
\midrule
Original        & 0.0142 & 0.0182 & 0.1392 & --- \\
Same-task swap  & 0.0310 & 0.4646 & 0.4036 & +189.9\% \\
Cross-task swap & 0.0341 & 0.6394 & 0.4838 & +247.6\% \\
Zero tokens     & 0.0289 & 0.4020 & 0.3907 & +180.7\% \\
\bottomrule
\end{tabular}
\end{table}

\paragraph{Results.}
Table~\ref{tab:token_dependency} reports MSE for the predicted DINOv3
feature deltas and depth-map deltas, together with the world-model loss defined in
Appendix~\ref{sec:loss_implementation}, before multiplication by its outer policy-loss
weight. Both same-task and cross-task swaps increase prediction error.
After averaging the three donors, the world-model loss increases for all
240 recipient clips under either swap. Same-task replacement already
raises this loss by 189.9\%, indicating that a shared task identity does
not substitute for the original sample-specific tokens. Cross-task
replacement produces a larger increase of 247.6\%.
These results show that the decoder relies on policy-generated
latent actions to predict future semantic and geometric changes.

\begin{figure}[!htb]
    \centering
    \includegraphics[width=\linewidth]{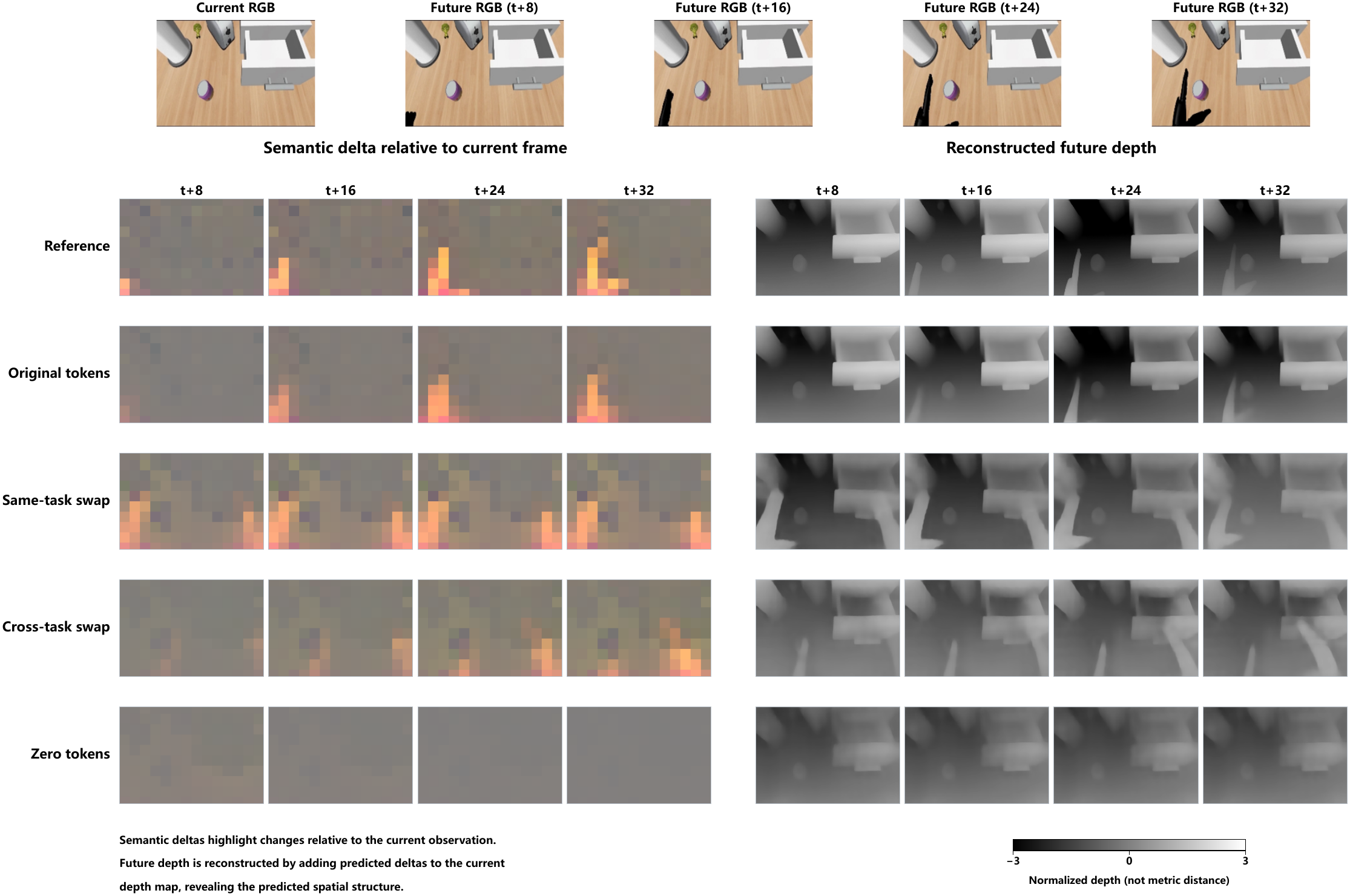}
    \caption{Decoder predictions under token replacement.
    Left: semantic-feature deltas relative to the current observation,
    displayed with a shared PCA projection and color scale across all
    conditions and future frames. Gray denotes zero projected change;
    colors do not represent fixed semantic categories.
    Right: future depth reconstructed by adding predicted depth deltas
    to the current depth map.
    Reference feature deltas and depth maps are derived from observed
    frames using DINOv3 and Depth Anything V2.}
    \label{fig:token_dependency}
\end{figure}

Figure~\ref{fig:token_dependency} shows one of six examples selected
before inspecting the intervention results. With the original tokens,
the predicted semantic changes and future geometry follow the reference
motion. Same-task and cross-task replacements instead introduce changes
in incorrect regions and distort the predicted geometry and temporal
progression, despite an unchanged current observation. With zero tokens,
the projected semantic deltas become nearly neutral gray, while the
reconstructed depth changes little across future steps and largely
retains the current scene layout. In this example, removing the latent
action information suppresses the predicted motion, whereas replacing
it with another sample's tokens produces mismatched changes.

\subsection{Action Probes of Frozen Latent Representations}
\label{sec:action_probes}

\paragraph{Data and targets.}
We test how readily recorded robot actions can be recovered from frozen
representations. The WALA LAM is taken from video pretraining, before
robot-policy training. For a fair comparison on RoboTwin and RoboCasa,
UniVLA~\citep{bu2025univlalearningacttaskcentric}
and Moto~\citep{chen2025motolatentmotiontoken} use LAM checkpoints
fine-tuned on RoboTwin and RoboCasa. The same checkpoints are used for
their encoder-attention visualizations in Fig.~\ref{fig:attention_vis}.
All LAMs remain frozen during probe feature extraction.
For RoboCasa-GR1-Tabletop, we sample 100 episodes per task and split them
into 80 training, 10 validation, and 10 test episodes. For RoboTwin, we
use 50 Clean and 50 Random episodes per task, with a 40/5/5 split within
each setting. For the official RoboDojo dataset~\citep{chen2026robodojo},
we use 100 episodes from each of 18 real-robot and 35 simulated task
groups, with an 80/10/10 split within each group.
Clean and Random data are pooled for the RoboTwin probe, and real and
simulated data are pooled for the RoboDojo probe.
We sample up to three non-overlapping action windows per episode and
repeat sampling and probe fitting with seeds 42, 43, and 44. All compared
representations use the same episodes and windows within each seed.
RoboCasa windows contain 32 actions and frames at offsets
$[0,8,16,24,32]$; RoboTwin and RoboDojo windows contain 50 actions and
frames at $[0,12,25,38,50]$. The probe predicts recorded action sequences,
using arm, hand, and waist actions throughout fitting and evaluation on
RoboCasa, and the bimanual action vectors on RoboTwin and RoboDojo.
No language, robot state, or wrist images are supplied.
WALA's LAM pretraining includes RoboCasa and RoboTwin training videos
(Appendix~\ref{sec:pretraining_data}), but neither the real nor simulated
RoboDojo videos. RoboDojo therefore tests the frozen WALA representation
on an unseen dataset, with a new probe fitted on its training split.

\begin{table}[!b]
\centering
\caption{Action decoding from frozen representations.
Test MSE is measured in standardized action coordinates and reported as
mean $\pm$ sample standard deviation over three seeds; lower is better.
``Current'' denotes current visual features. For UniVLA and Moto latent actions,
``continuous'' denotes encoder outputs before vector quantization;
``quantized'' denotes the corresponding codebook embeddings.}
\label{tab:action_probes}
\small
\renewcommand{\arraystretch}{1.14}
\setlength{\tabcolsep}{4.5pt}
\begin{tabular}{lccc}
\toprule
Probe input
& \makecell{RoboCasa-GR1-\\Tabletop}
& \makecell{RoboTwin\\Clean + Random}
& \makecell{RoboDojo\\Real + Sim} \\
\midrule
Training action mean (no visual input)
& $0.958 \pm 0.060$ & $0.996 \pm 0.039$ & $1.005 \pm 0.014$ \\
Current visual features
& $0.657 \pm 0.078$ & $0.587 \pm 0.032$ & $0.541 \pm 0.005$ \\
Visual feature sequence
& $0.605 \pm 0.083$ & $0.477 \pm 0.052$ & $0.361 \pm 0.013$ \\
\midrule
Current + UniVLA latent actions (quantized)
& $0.637 \pm 0.079$ & $0.526 \pm 0.034$ & $0.493 \pm 0.006$ \\
UniVLA latent actions (quantized)
& $0.867 \pm 0.063$ & $0.787 \pm 0.036$ & $0.828 \pm 0.012$ \\
Current + UniVLA latent actions (continuous)
& $0.614 \pm 0.075$ & $0.432 \pm 0.040$ & $0.383 \pm 0.004$ \\
UniVLA latent actions (continuous)
& $0.688 \pm 0.067$ & $0.492 \pm 0.045$ & $0.500 \pm 0.006$ \\
\midrule
Current + Moto latent actions (quantized)
& $0.660 \pm 0.076$ & $0.448 \pm 0.034$ & $0.387 \pm 0.004$ \\
Moto latent actions (quantized)
& $0.817 \pm 0.063$ & $0.541 \pm 0.038$ & $0.538 \pm 0.015$ \\
Current + Moto latent actions (continuous)
& $0.650 \pm 0.067$ & $0.418 \pm 0.036$ & $0.339 \pm 0.005$ \\
Moto latent actions (continuous)
& $0.747 \pm 0.057$ & $0.472 \pm 0.039$ & $0.423 \pm 0.009$ \\
\midrule
Current + WALA latent actions
& $\mathbf{0.541 \pm 0.070}$ & $0.299 \pm 0.034$ & $\mathbf{0.308 \pm 0.006}$ \\
WALA latent actions
& $0.550 \pm 0.061$ & $\mathbf{0.287 \pm 0.036}$ & $0.358 \pm 0.011$ \\
\bottomrule
\end{tabular}
\end{table}

\paragraph{Representations and probe fitting.}
Current visual features concatenate DINOv3 features and normalized depth
from the primary view. The visual feature sequence combines this
current feature vector with four future feature deltas,
$[v_t;\,\{v_{t+k}-v_t\}_{k\in\mathcal{K}}]$, where $\mathcal{K}$ is the set
of four future-frame offsets and $v$ includes both semantic and depth features.

For UniVLA and Moto, we concatenate representations from four
consecutive observation pairs spanning the same five-frame clip.

Each feature group is projected by PCA using the same 4,096 training
windows, retaining up to 512 nondegenerate components, followed by
scaling each projected coordinate by its training standard deviation.
We fit linear ridge probes with an unpenalized intercept,
selecting regularization on the full action targets in validation data from
$\{0,10^{-10},10^{-9},\ldots,10^1\}$. Action coordinates are standardized
using training-set statistics. Table~\ref{tab:action_probes} reports test
MSE averaged over action coordinates, timesteps, and windows, followed by
the mean and sample standard deviation across the three seeds. The
no-visual baseline predicts the training-set mean action at every step.

\paragraph{Results.}
Combining current visual features with WALA latent actions yields lower
mean error than the visual baselines and both representations of UniVLA and Moto on
all three datasets. It achieves the lowest error on RoboCasa (0.541) and
RoboDojo (0.308), while WALA latent actions alone perform best on
RoboTwin (0.287). Without current visual features, WALA also outperforms
both representations of UniVLA and Moto on all three datasets. Together,
these comparisons show that WALA's semantic-geometric latent actions
retain information that more readily supports linear decoding of robot
actions, including on RoboDojo videos not used in WALA pretraining.

The benefit of current visual features depends on the dataset. WALA
latent actions alone are competitive on RoboCasa and strongest on
RoboTwin, whereas adding current visual features reduces RoboDojo error
from 0.358 to 0.308. Thus, the latent actions carry substantial
action-related information, while current semantic-geometric observations
can provide complementary context.

For both UniVLA and Moto, continuous latent actions consistently yield lower error than
their quantized counterparts, both with and without current visual
features. This suggests that vector quantization may discard some
action-related information that is accessible to a linear probe.

\paragraph{Clean-to-Random transfer.}
We further evaluate whether an action probe fitted on RoboTwin Clean
transfers to Random. Both settings contain the same manipulation tasks.
Clean uses plain backgrounds and uncluttered tabletops, whereas Random
varies background and tabletop textures, adds randomly placed distractor
objects, and randomizes lighting and tabletop
height~\citep{chen2025robotwin20scalabledata}.
For each task, we use 45 Clean episodes for fitting,
5 Clean episodes for hyperparameter selection, and 50 Random episodes
for testing. We retain the action windows and targets described above
and repeat the experiment with seeds 42, 43, and 44. PCA, feature scaling,
and action normalization are fitted using only the Clean training split.
We report the same standardized action MSE as
Table~\ref{tab:action_probes}, using these Clean-training statistics.
The visual baselines concatenate current features with either absolute
future features or their differences from the current features,
$[v_t;\,\{v_{t+k}\}_{k\in\mathcal{K}}]$ and
$[v_t;\,\{v_{t+k}-v_t\}_{k\in\mathcal{K}}]$, respectively.
We use the same per-group PCA procedure as above. UniVLA and Moto latent
actions in this experiment are taken before vector quantization.

\begin{table}[htbp]
\centering
\caption{Action-probe transfer from RoboTwin Clean to Random.
Standardized action MSE is reported as mean $\pm$ sample standard
deviation over three seeds; lower is better.}
\label{tab:clean_random_probe}
\small
\renewcommand{\arraystretch}{1.14}
\setlength{\tabcolsep}{8pt}
\begin{tabular}{lc}
\toprule
Probe input & Standardized action MSE $\downarrow$ \\
\midrule
Absolute visual feature sequence & $0.9408 \pm 0.0782$ \\
Delta visual feature sequence & $0.9219 \pm 0.0654$ \\
UniVLA latent actions & $0.7173 \pm 0.0249$ \\
Moto latent actions & $2.7119 \pm 0.1852$ \\
WALA latent actions & $\mathbf{0.4958 \pm 0.0248}$ \\
\bottomrule
\end{tabular}
\end{table}

Table~\ref{tab:clean_random_probe} shows the lowest error with WALA latent
actions, outperforming UniVLA (0.7173) and Moto (2.7119).
This comparison favors WALA's representation for transferring action
decoding across visual settings, beyond fitting actions within a shared setting.
Emphasizing action-relevant semantic and geometric changes may help
retain information useful across settings while reducing the probe's
dependence on setting-specific appearance. Visual deltas also yield a
slightly lower mean error than absolute future features.

\subsection{Effect of Decoder Supervision and Fine-Tuning}
\label{sec:decoder_adaptation}

Matching a frozen LAM encoder's outputs transfers pretrained knowledge to the policy. We examine whether additional guidance from the LAM decoder improves future prediction from policy-generated latent actions on downstream interactions. Using RoboDojo videos not seen during LAM pretraining, we compare training without decoder supervision, supervision through a frozen decoder, and joint decoder fine-tuning.

\paragraph{Protocol.}
We pool real and simulated RoboDojo data from 53 task--embodiment groups,
using 100 trajectories per group. For each group, we use 80 trajectories
for training and the remaining 20 for evaluation. Training windows start every four frames
and contain a 50-step action chunk, with observations at offsets
$[0,12,25,38,50]$. Each setting is trained for one epoch on 16 Alibaba
Cloud PPUs with a global batch size of 128. Seeds 42, 43, and 44 require
5,021, 5,030, and 5,025 optimizer updates, respectively. We evaluate the
final checkpoint from each run. Within each seed, all three settings share the same data splits and policy initialization. All start from
Qwen3-VL-4B-Instruct and the pretrained WALA LAM; the LAM encoder and visual
encoders remain frozen. The learning rates are $10^{-5}$ for the policy
backbone and trainable decoder parameters, and $10^{-4}$ for the action
head and adapters.

\paragraph{Compared settings.}
All three settings retain robot action supervision and latent-action
distillation. Setting A omits decoder prediction supervision. Setting B
adds the semantic-geometric prediction loss through a frozen decoder,
allowing gradients to update the policy and decoder adapter. Setting C
also updates the decoder-specific parameters; shared encoder--decoder
components remain frozen. For prediction evaluation, A sends policy tokens
through its trained distillation adapter to the pretrained decoder,
whereas B and C use their trained decoder adapters.

\paragraph{Metrics.}
We report semantic-feature delta MSE and depth-map delta MSE on real and simulated RoboDojo test trajectories. Table~\ref{tab:decoder_adaptation} reports the mean and standard deviation across three random seeds.

\begin{table}[htbp]
\centering
\caption{Effect of decoder supervision and fine-tuning on real and simulated RoboDojo data. Semantic and depth delta MSE are reported as mean $\pm$ standard deviation across three random seeds after one training epoch. All settings retain robot action supervision and latent-action distillation.}
\label{tab:decoder_adaptation}
\small
\renewcommand{\arraystretch}{1.12}
\setlength{\tabcolsep}{9pt}
\begin{tabular}{lcc}
\toprule
Setting & \makecell{Semantic delta\\MSE $\downarrow$}
& \makecell{Depth delta\\MSE $\downarrow$} \\
\midrule
A: No decoder supervision & $0.03365 \pm 0.00101$ & $0.7018 \pm 0.0028$ \\
B: Frozen decoder        & $0.02167 \pm 0.00020$ & $0.3498 \pm 0.0041$ \\
C: Trainable decoder     & $\mathbf{0.01912 \pm 0.00002}$ & $\mathbf{0.3355 \pm 0.0037}$ \\
\bottomrule
\end{tabular}
\end{table}

\paragraph{Results.}
Adding prediction supervision through a frozen decoder reduces semantic
delta MSE by 35.6\% and depth delta MSE by 50.2\% relative to A.
Prediction therefore improves even without updating the
decoder, supporting the value of decoder-based prediction
supervision in addition to latent-action distillation. Jointly
adapting the decoder further reduces the two errors by 11.8\% and 4.1\%,
with a larger relative improvement in semantic prediction. These results
support using the pretrained decoder to guide downstream learning and
adapting its predictions to the downstream interaction data.

\subsection{Action Readout from All Backbone Tokens}
\label{sec:all_tokens_readout}

\paragraph{Protocol.}
We examine whether the action head benefits from access to all final-layer
VLM tokens rather than only the supervised latent action tokens. The all-token
variant uses a single cross-attention resampler with 32 learned queries and
8 attention heads to aggregate the variable-length backbone sequence into
32 features for the original MLP action head. Encoder distillation and
decoder prediction still supervise the original latent action tokens.
The data, loss functions, and action-chunk length follow the standard
RoboCasa configuration; the resampler introduces additional parameters.

\begin{figure}[htbp]
    \centering
    \includegraphics[width=0.80\linewidth]{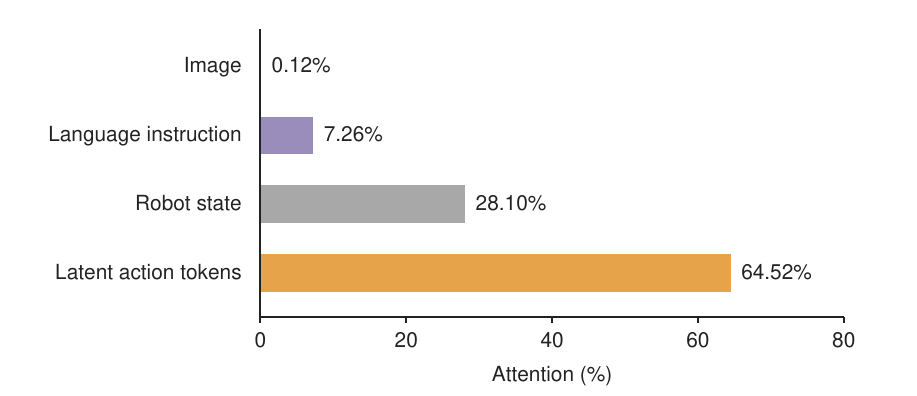}
    \caption{Action-readout attention in the all-token variant on RoboCasa.
    Statistics cover 100 samples per task across 24 tasks. Latent action
    tokens receive the highest average attention.}
    \label{fig:all_tokens_attention}
\end{figure}

\paragraph{Manipulation performance.}
The all-token variant achieves 73.3\% average success, compared with
75.2\% for WALA's latent-token-only action readout, a difference of
1.9 percentage points. Broader access to backbone features therefore
does not improve on the reported WALA result. This comparison addresses
the design question raised in the Introduction: latent actions can form
the direct interface to robot control rather than serve only as an
auxiliary prediction target. WALA concentrates interaction-prediction
and action-generation supervision on this interface, so the action head
reads a representation explicitly shaped by both objectives.

\paragraph{Attention analysis.}
For each task, we sample 100 distinct training trajectories and one random
time step per trajectory, yielding 2,400 observations. The diagnostic inputs
include the image, language instruction, and robot state. We average the
resampler's attention over its 8 heads and 32 queries, divide each category's
attention by its token count, normalize across the four categories, and
average over samples; formatting tokens are excluded from the statistics.
Figure~\ref{fig:all_tokens_attention} shows that latent action tokens receive
the highest average attention (64.52\%), followed by robot state (28.10\%),
language instructions (7.26\%), and image tokens (0.12\%). Latent action
tokens rank first in all 24 tasks. Even when the action head can access
the full backbone sequence, its readout preferentially attends to the
world- and action-supervised latent representation. Together with the
success-rate comparison, this pattern supports using latent action tokens
as the action head's sole input: access to the broader sequence neither
displaces their prominent role nor yields a higher reported success rate.
The substantial attention to robot state is consistent with its
role in relating the robot's current configuration to subsequent
movements. Through contextual processing in the backbone, latent action tokens
can also aggregate image, language, and robot-state information into
a representation jointly shaped by world and action supervision.
Directly decoding actions from these tokens thus provides an efficient
and lightweight approach to robot control while maintaining strong
manipulation performance.

\end{document}

%% file: math_commands.tex
\usepackage{amsmath,amsfonts,bm}

\def\eqref#1{equation~\ref{#1}}

\def\1{\bm{1}}

\DeclareMathAlphabet{\mathsfit}{\encodingdefault}{\sfdefault}{m}{sl}
\SetMathAlphabet{\mathsfit}{bold}{\encodingdefault}{\sfdefault}{bx}{n}

